\pdfoutput=1
\documentclass[11pt]{article}

\usepackage[final]{acl}
\usepackage{amsmath}
\usepackage{booktabs}
\usepackage{multirow}
\usepackage{authblk}
\usepackage{longtable}

\usepackage{times}
\usepackage{latexsym}
\usepackage{listings}
\usepackage{xcolor}
\lstdefinestyle{promptstyle}{
  basicstyle=\ttfamily\footnotesize,
  breaklines=true,
  columns=fullflexible,
  backgroundcolor=\color{gray!10},
  frame=single,
  showstringspaces=false,
  escapeinside={(*@}{@*)}
}

\usepackage[T1]{fontenc}
\usepackage[utf8]{inputenc}

\usepackage{microtype}

\usepackage{inconsolata}

\usepackage{graphicx}

\usepackage{algorithm}
\usepackage{algorithmic}
\usepackage[most]{tcolorbox}
\usepackage[dvipsnames,svgnames,x11names]{xcolor}
\definecolor{mycolor}{HTML}{D5E8D4}
\definecolor{mycolor2}{HTML}{f2f8f2}
\definecolor{mycolor3}{HTML}{DAE8FC}
\definecolor{mycolor4}{HTML}{F4F8FE}
\definecolor{mycolor5}{HTML}{FFCCCC}
\definecolor{mycolor6}{HTML}{FAE5E5}

\usepackage[T1]{fontenc}
\usepackage{fancyvrb}
\usepackage{inconsolata}

\title{CATCH: A Novel Data Synthesis Framework for High Therapy Fidelity and Memory-Driven Planning Chain of Thought in AI Counseling}

\author{
  \vspace{-8pt}  % 减小下一行的间距
  \textbf{Mingyu Chen}\textsuperscript{1}\thanks{Equal contribution.}, 
  \textbf{Jingkai Lin}\textsuperscript{1}\textsuperscript{$\ast$}, 
  \textbf{Zhaojie Chu}\textsuperscript{1}, \\ 
  \textbf{Xiaofen Xing}\textsuperscript{1}\thanks{Corresponding author.}, 
  \textbf{Yirong Chen}\textsuperscript{1}, 
  \textbf{Xiangmin Xu}\textsuperscript{2,1}\textsuperscript{†} \\
  \textsuperscript{1}School of EE., South China University of Technology, Guangzhou, China \\
  \textsuperscript{2}Foshan University, Foshan, China \\
  \{eemingyuchen, eelinjingkai, 202210182056, eeyirongchen\}@mail.scut.edu.cn, \\
  \{xfxing, xmxu\}@scut.edu.cn
}

\begin{document}
\maketitle
\begin{abstract}
Recently, advancements in AI counseling based on large language models have shown significant progress. However, existing studies employ a one-time generation approach to synthesize multi-turn dialogue samples, resulting in low therapy fidelity and failing to capture the decision-making rationale behind each response. In this work, we propose CATCH, a novel data synthesis framework designed to address these challenges. Specifically, to improve therapy fidelity, we introduce the Progressive Dialogue Synthesis strategy, which extracts goals, resources, and solutions from a client’s self-report, organizes them into structured outlines, and then incrementally generates stage-aligned counseling dialogues. To capture decision-making rationale behind each response, we propose the Memory-Driven Dynamic Planning thinking pattern that integrates memory enhancement, global planning, and strategy reasoning; a collaborative multi-agent optimizer then leverages MDP to attach explicit chain-of-thought to each dialogue turn. Extensive experiments and human evaluations demonstrate that CATCH significantly enhances fidelity and logical coherence in AI counseling.\footnote{https://github.com/scutcyr/SoulChat-R1}
\end{abstract}

% origin 主要修改了里边比较专业术语些的描述
% Recently, advancements in AI counseling based on large language models have shown significant progress. However, existing studies employ a one-time generation approach to synthesize multi-turn dialogue samples, resulting in low therapy fidelity and a lack of deeper understanding of the therapeutic logic underlying each response turn. In this work, we propose a novel data synthesis framework, CATCH, designed to address the challenges above. Specifically, to improve therapy fidelity, we introduce the Progressive Dialogue Synthesis strategy to systematically derive key elements from the client's self-report and organize them into structured outlines, thereby generating a counseling dialogue dataset aligned with therapeutic intervention principles. To enhance therapy logic, we propose the Memory-Driven Dynamic Planning thinking pattern, which clarifies decision-making motivations for each dialogue turn. This pattern incorporates memory enhancement, global planning, and strategy reasoning, enabling the development of a collaborative multi-agent iterative optimization method that synthesizes complete chain of thought in counseling dialogues. Extensive experiments demonstrate that CATCH significantly enhances the fidelity and logical coherence in AI counseling.

\section{Introduction}
% origin
% Mental health issues, which are becoming increasingly prominent \cite{santomauro2021global}, have received significant attention. Counseling serves as an important means of alleviating mental health issues. Professional mental health counselors are scarce and costly to serve, making it difficult to meet the needs of a large number of clients. Recently, the language understanding and generation capabilities of Large Language Models (LLMs) have been significantly improved \cite{openai2024gpt4technicalreport,guo2025deepseek, yang2025qwen3technicalreport}, which provides new solutions and support tools for the field of mental health counseling \cite{hua2024large, xiao-etal-2024-healme,xie-etal-2025-psydt,lee-etal-2024-cactus,xu2025autocbt}. To enhance the expertise of LLMs in the field of mental health counseling, a large-scale dataset of multi-turn counseling dialogues from the real world is essential.

% new 
Mental health issues are becoming increasingly prominent worldwide \cite{santomauro2021global}, highlighting counseling as a crucial intervention method. However, due to the scarcity and high costs associated with professional counselors, a significant gap persists between service availability and demand. Recent advancements in Large Language Models (LLMs) \cite{openai2024gpt4technicalreport,guo2025deepseek, yang2025qwen3technicalreport} offer promising opportunities to bridge this gap, providing novel tools for mental health counseling \cite{hua2024large, xiao-etal-2024-healme,xie-etal-2025-psydt,lee-etal-2024-cactus,xu2025autocbt}. To effectively leverage LLMs in mental health counseling, large-scale, high-quality datasets of multi-turn counseling dialogues are essential.

% origin
% Although these studies synthesize large-scale datasets of multi-turn counseling dialogues, they use a one-time generation approach to directly synthesize multi-turn dialogue samples \cite{qiu-etal-2024-smile,zhang-etal-2024-cpsycoun,xie-etal-2025-psydt,lee-etal-2024-cactus}. These synthetic datasets pose challenges for AI counselors in effectively applying the fundamental principles of the therapy at each stage, resulting in \textbf{low therapy fidelity} \cite{waltz1993testing, bellg2004enhancing, perepletchikova2005treatment, sanetti2021treatment} in the synthesized dialogues. Additionally, these dialogue data lack explicit constraints on the decision-making motivations for each response turn. As a result, AI counseling models developed from these synthetic datasets tend to superficially mimic the content of the responses, \textbf{lacking a deeper understanding of the therapy logic} underlying each response turn. These challenges cause AI counseling models to exhibit only limited ability to empathize and provide basic advice and guidance in each turn of response, but to suffer from \textbf{therapy drift} \cite{waller2009evidence, waller2016therapist} in multi-turn conversations.

% new  applying the fundamental principles of the therapy at each stage 改为adheres to the structured stages of a given therapy 以及补充therapy fidelity和therapy drift的引用和解释
Although recent methods have enabled the synthesis of large-scale multi-turn counseling dialogue datasets, they typically adopt a \textbf{one-time generation approach}\cite{qiu-etal-2024-smile,zhang-etal-2024-cpsycoun,lee-etal-2024-cactus,xie-etal-2025-psydt}, where a Large Language Model (LLM) produces the entire dialogue in a single inference step. This approach offers limited control over the therapeutic process, making it difficult to ensure that the dialogue adheres to the structured stages of a given therapy, resulting in low \textbf{therapy fidelity}\cite{waltz1993testing, bellg2004enhancing, perepletchikova2005treatment, sanetti2021treatment}—the degree to which an intervention aligns with established therapeutic guidelines. Furthermore, these datasets generally lack explicit chain-of-thought (CoT) \cite{wei2022chain} that expose the decision-making rationale behind each counselor response. Consequently, models trained on such data tend to imitate surface-level conversational patterns rather than grasping the underlying therapeutic logic. These challenges cause AI counseling models to suffer from \textbf{therapy drift} \cite{waller2009evidence, waller2016therapist}, where the dialogue progressively deviates from the intended therapeutic trajectory.

In this work, we introduce CATCH, a \textbf{C}-ollaborative multi-\textbf{A}-gent-based framework for \textbf{T}-urn-level \textbf{C}-hain-of-Thought and \textbf{H}-igh-fidelity counseling dialogues synthesis, thereby mitigating therapy drift. To enhance therapy fidelity, CATCH adopts a Progressive Dialogue Synthesis (PDS) strategy, which incrementally generates dialogues stage by stage, aligning with structured therapeutic intervention protocols. Starting from client self-reports, PDS identifies key elements—counseling goals, resources, and solutions—and organizes them into outlines. These outlines, combined with the client’s personality traits, guide LLMs to synthesize the dialogue for each therapeutic stage. To explicitly capture the decision-making rationale behind each counselor response, we introduce a Memory-driven Dynamic Planning (MDP) thinking pattern that incorporates memory enhancement, global planning, and strategy reasoning. Built upon MDP, we develop a collaborative multi-agent approach to generate MDP CoT. To evaluate CATCH's effectiveness, we developed 8B and 14B scale models using dialogue datasets enriched with MDP CoT. Extensive experiments and human evaluations demonstrate that CATCH improves the fidelity and logical coherence of AI counseling, effectively mitigating therapy drift.

% origin
% The contributions of this work are as follows:
% \begin{itemize}
%   \item We propose a PDS strategy to generate a counseling dialogue dataset aligned with therapy intervention principles. By utilizing the client's self-report, PDS systematically derives key elements and organizes them into structured outlines, enhancing the quality of multi-turn dialogues in AI counseling.
%   \item We design an MDP thinking pattern to clarify decision-making motivations for each dialogue turn. This pattern incorporates memory enhancement, global planning, and strategy reasoning, enabling the development of a collaborative multi-agent iterative optimization method that synthesizes complete CoT in counseling dialogues.
%   \item Extensive experiments demonstrate that CATCH enhances the fidelity and logical coherence of AI counseling while effectively mitigating therapy drift.
% \end{itemize}

% new 将前两个贡献的表述改了一下，和前文保持一致
The contributions of this work are as follows:
\begin{itemize}
    \item We propose the Progressive Dialogue Synthesis (PDS) strategy. Unlike one-time generation approach, PDS incrementally synthesizes dialogues stage by stage, aligning with structured therapeutic protocols and deriving key elements from client self-reports to guide LLM-based content generation for each therapeutic stage.
    \item We introduce the Memory-driven Dynamic Planning (MDP) thinking pattern—comprising memory enhancement, global planning, and strategy reasoning—and devise a collaborative multi-agent method to produce explicit turn-level chain-of-thought, making the counselor’s decision-making rationale transparent.
    \item Extensive experiments and human evaluations demonstrate that CATCH enhances the fidelity and logical coherence of AI counseling while effectively mitigating therapy drift.
\end{itemize}

\begin{figure*}[ht]
 \centering
  \includegraphics[width=4.8 in]{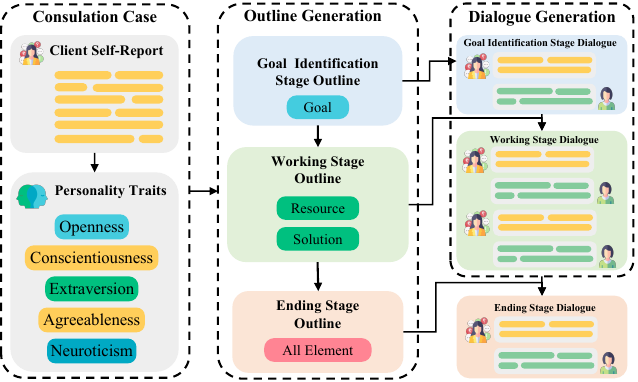} 
  \caption {The Progressive Dialogue Synthesis (PDS) strategy for high-fidelity dialogue generation. PDS translates therapeutic principles into a structured process: from client self-reports and personality traits, it derives goals, resources, and solutions, forming a stage-guided outline that ensures fidelity in the incremental generation of counseling dialogues.}
  \label{fig:PDS_strategy}
\end{figure*}

\section{Related Work}

\subsection{Psychological Models}

With the growing application of LLMs, the mental health domain has seen a surge in LLMs-based approaches tailored for counseling scenarios. Early efforts primarily focused on data collection and construction. For example, SoulChat \cite{chen-etal-2023-soulchat} built a large-scale corpus with over 2 million dialogue samples, while MindChat \cite{MindChat} curated approximately 200,000 multi-turn counseling dialogues through manual cleaning, providing a valuable foundation for downstream modeling.

To improve the modeling of specific intervention strategies, some studies have explored the integration of structured therapy procedures\cite{xu2025autocbt,xiao-etal-2024-healme}. AutoCBT \cite{xu2025autocbt} developed a single-turn multi-agent system for cognitive behavioral therapy (CBT). Healme \cite{xiao-etal-2024-healme} decomposed the cognitive restructuring process of CBT into three actionable dialogue stages, using prompts to guide the LLMs step-by-step in assisting the client. However, these systems typically handle only a limited number of turns and are insufficient to capture the complexity of real-world multi-turn counseling interactions.

Due to the scarcity of authentic multi-turn counseling data, some studies have investigated the synthesis of such dialogues. SMILE \cite{qiu-etal-2024-smile} expands single-turn counseling interactions to create multi-turn dialogues. CpsyCoun \cite{zhang-etal-2024-cpsycoun} reconstructs multi-turn dialogues based on counseling reports. Other works target therapy-specific dialogue synthesis: Cactus \cite{lee-etal-2024-cactus} leverages CBT plans to guide the generation of multi-turn dialogues, and PsyDT \cite{xie-etal-2025-psydt} constructs a twin dialogue dataset of counselors by simulating the therapeutic language style of counselors. Although these studies synthesize large-scale datasets of multi-turn counseling dialogues, they use a one-time generation approach to directly synthesize multi-turn dialogue samples. These synthetic datasets pose challenges for AI counselors in effectively applying the fundamental principles of therapy at each stage, resulting in low therapy fidelity for the synthesized dialogues.

\subsection{Structured Psychotherapy}

% origin 这里的改动主要是缩写和分点
% Structured psychotherapy refers to therapeutic approaches conducted under clear theoretical frameworks with standardized procedures and predefined goals. These methods emphasize goal orientation, procedural clarity, and short-term effectiveness, making them suitable for evaluation and replication. CBT \cite{beck2020cognitive} is the most widely applied structured therapy. It focuses on identifying and restructuring negative thought patterns and typically includes 8–16 sessions with clear, goal-driven steps. Dialectical Behavior Therapy (DBT) \cite{robins2011dialectical} extends CBT with mindfulness and emotion regulation modules. It features modular training and clear targets, making it effective for disorders involving emotional dysregulation. Solution-Focused Brief Therapy (SFBT) \cite{zak2024effectiveness} aims to construct solutions rather than analyze problems. It is typically completed within 4–8 sessions and emphasizes the activation of client resources. Single-Session Therapy (SST) \cite{kim2023effectiveness} treats each session as a complete intervention, focusing on the client’s most urgent concern. It follows a structured flow involving problem focus, goal setting, resource activation, and action planning.

Structured psychotherapy refers to therapeutic approaches guided by clear theoretical frameworks, standardized procedures, and predefined objectives. These therapies are characterized by goal-oriented practices, clear procedures, and a focus on short-term effectiveness, which makes them easier to evaluate and replicate \cite{chambless1998defining, lambert2013bergin}. Prominent examples include:

\noindent \textbf{Cognitive Behavioral Therapy} (CBT) \cite{beck2020cognitive}, the most widely applied structured therapy, aims to identify and restructure negative thought patterns. It usually involves 8–16 sessions with clearly defined steps and goals.
  
\noindent \textbf{Dialectical Behavior Therapy} (DBT) \cite{robins2011dialectical} builds upon CBT by adding mindfulness and emotion regulation techniques. It employs modular training and explicit therapeutic targets, making it particularly suitable for disorders characterized by emotional instability.
  
\noindent \textbf{Solution-Focused Brief Therapy} (SFBT) \cite{zak2024effectiveness} emphasizes constructing solutions instead of analyzing problems, typically completed within 4–8 sessions. It focuses on activating the client’s internal resources.
  
\noindent \textbf{Single-Session Therapy} (SST) \cite{kim2023effectiveness} considers each session as an independent intervention addressing the client's most pressing issue. It follows a structured procedure: setting the counseling goal, activating resources, and planning actions.

\begin{figure*}
    \centering
    \includegraphics[width=5.5 in]{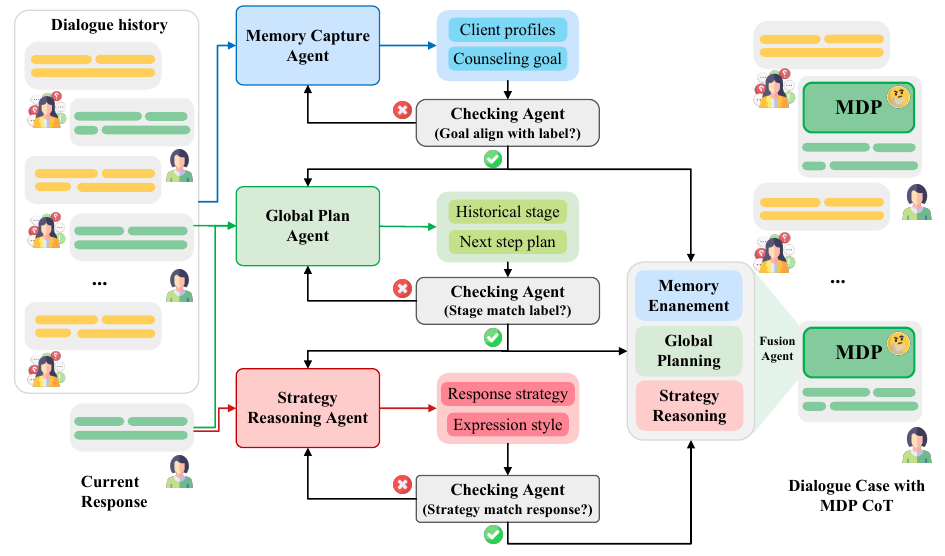}
    \caption{The Memory-Driven Dynamic Planning (MDP) Chain-of-Thought (CoT) synthesis pipeline. The framework employs a collaborative multi-agent system to generate structured counseling reasoning: the Memory Capture Agent summarizes and extracts key information from the dialogue history, the Global Plan Agent determines therapeutic stage and progression, and the Strategy Reasoning Agent refines response strategies. Each step is verified by a Checking Agent for consistency, and the Fusion Agent integrates validated components into a coherent, first-person thought process.}
    \label{fig:mdp_pipeline}
\end{figure*}

\section{Methodology}

% 把therapy intervention principles删除掉 原先是 to generate a counseling dialogue datase aligned with therapy intervention principles
In this work, we present a comprehensive introduction to the proposed data synthesis framework, CATCH, designed for structured psychotherapy. This work uses SST as a specific therapy. In Section ~\ref{Synthesis_dataset}, we describe the \textbf{Progressive Dialogue Synthesis} (PDS) to generate a counseling dialogue dataset with high therapy fidelity. Section ~\ref{Thinking_pattern} introduces the generation of the \textbf{Memory-driven Dynamic Planning} (MDP) Chain-of-Thought(CoT). Section ~\ref{develop_LLMs} focuses on developing a LLMs-based AI counselor utilizing the synthesized dataset enriched with the MDP CoT.

\subsection{PDS Strategy}
\label{Synthesis_dataset}
 
% therapy intervention principles 修改成了 structured therapeutic protocols
Existing work synthesizing datasets of counseling dialogues is challenged by low therapy fidelity \cite{qiu-etal-2024-smile,zhang-etal-2024-cpsycoun,xie-etal-2025-psydt,lee-etal-2024-cactus}. We propose the PDS strategy to generate a counseling dialogue dataset aligned with structured therapeutic protocols. Detailed prompts used in PDS are provided in Appendix~\ref{PDS Prompts}

\subsubsection{Consultation Case Acquisition}

To ensure the authenticity and effectiveness of the synthesized counseling dialogues, we collected posts describing mental health from Yixinli \footnote{https://www.xinli001.com/}. Each post contains the client's self-report text, which serves as the initial information for subsequent dialogue creation, as shown in Figure~\ref{fig:PDS_strategy} (left). Following PysDT \cite{xie-etal-2025-psydt}, we extracted the client's Big Five personality traits from the self-report text as supporting information.

\subsubsection{Outline Generation}
\label{subsubsection:outline_generation}

Unlike the existing dialogue generation approach, we use the PDS strategy to enhance the therapy fidelity. As shown in Figure~\ref{fig:PDS_strategy} (middle), our proposed PDS strategy requires the sequential performance of a goal identification stage, a working stage, and an ending stage. The elements of counseling dialogue creation are obtained. Then, based on these elements, we utilize LLM to generate an outline of the dialogue for each therapy stage. Terminology explanations are provided in Appendix ~\ref{Terminology_explanation}.

% origin 没有说明goal如何获取 以及对于目标的要求说得比较模糊 unique, specific, feasible, positive, evaluable, and focused on a psychological difficulty rather than a real problem --->  unique, specific, feasible, positive, evaluable, and focused on the client's internal experiences and responses rather than on changing external circumstances or the actions of others. 
% \textit{Element Acquisition:} Based on the client's self-report and the Big Five personality traits, we obtain a counseling goal element \(\mathcal{G}\) that meets the assessment criteria for SST therapy and is unique, specific, feasible, positive, evaluable, and focused on a psychological difficulty rather than a real problem. \textit{Outline Generation:} The counseling dialogue begins with the counselor establishing an initial trusting relationship with the client, progressively explores the client's current distress, and ultimately clarifies and confirms the counseling goal \(\mathcal{G}\) at the ending stage. Detailed prompts see Appendix \ref{target_generate_prompt}.
\noindent \textbf{Goal Identification Stage.}
\textit{Element Acquisition:} Based on the client's self-report and the Big Five personality traits, we use an LLM to obtain a counseling goal \(\mathcal{G}\) that meets the assessment criteria for SST therapy and is unique, specific, feasible, positive, evaluable, and focused on the client's internal experiences and responses rather than on changing external circumstances or the actions of others. \cite{de2021more}. \textit{Outline Generation:} The counseling dialogue begins with the counselor establishing an initial trusting relationship with the client, progressively explores the client's current distress, and ultimately clarifies and confirms the counseling goal \(\mathcal{G}\) at the ending stage. 

\noindent \textbf{Working Stage.}
\textit{Element Acquisition:} Combining the client's self-report and the counseling goal \(\mathcal{G}\), we employ the LLMs to explore multiple resources \(\mathcal{R}\) related to the client's demands, covering both negative resource \(\mathcal{R}_{N}\) and positive resource \(\mathcal{R}_{P}\). Based on the explored resources \(\mathcal{R}\) , the \(\mathcal{R}_{P}\) is used to generate solutions \(\mathcal{S}\). \textit{Outline Generation:} The counselor conducts a resource exploration around the counseling goal \(\mathcal{G}\). After several counseling explorations, the \(\mathcal{R}_{N}\) and \(\mathcal{R}_{P}\) are obtained. The \(\mathcal{R}_{P}\) is transformed into solutions \(\mathcal{S}\). 

\noindent \textbf{Ending Stage.}
\textit{Element Integration:}  All of the core elements from the first two stages are integrated, including consulting goals \(\mathcal{G}\), negative resources \(\mathcal{R}_{N}\), positive resources \(\mathcal{R}_{P}\), and solutions \(\mathcal{S}\). \textit{Outline Generation:} Under the guidance of the counselor, the client accepts and is willing to try to apply the solutions \(\mathcal{S}\). Moreover, the counselor summarizes the counseling process and assists the client in reinforcing the positive changes made.

\subsubsection{Dialogue Generation}

As shown in Figure~\ref{fig:PDS_strategy}(right), based on the dialogue outline (described in Section~\ref{subsubsection:outline_generation}), we utilized the client's personality traits and the SST knowledge to inform both the client's and counselor's language expression. The LLMs generate the consultation dialogue at each stage and attach the stage status label to each turn of responses. The dialogues for each subsequent stage were built upon the content generated in the previous stage, ultimately stitching together all stages to create a comprehensive sample of a multi-round counseling dialogue.

\subsubsection{Data Filtering}

To ensure the quality of the dialogue samples, we filtered out instances with abnormal turns and disordered statements. DeepSeek-R1\cite{guo2025deepseek} is employed to assess the therapeutic structure and goal focus of the dialogues, retaining only the data that passed this evaluation through a majority voting process. Additionally, the LLMs produce statements containing excessive technical jargon due to the provision of SST knowledge. Thus, we implement keyword matching to identify statements with an overabundance of specialized terms, again using DeepSeek-R1 for assessment. If a statement contained too many technical terms, we would revise it while preserving its original meaning. 

\subsection{MDP Thinking Pattern}
\label{Thinking_pattern}
% 3.2 Memory-Driven Dynamic Plan Thinking Pattern

\subsubsection{MDP Thinking Pattern Design}
\label{Pattern_design}

Case conceptualization \cite{kuyken2011collaborative} is the process through which the counselor systematically understands, integrates, and interprets the client's issues, providing guidance for subsequent counseling sessions. In this process, the counselor considers various factors, such as the client's personal information, background, and psychological state, to construct a comprehensive and in-depth case model. Based on this theory, we designed the MDP expert thinking pattern by simulating and modeling the case conceptualization process, aiming to help the model better understand the inherent therapeutic logic in multi-turn counseling dialogues. The thinking pattern contains Memory Enhancement, Global Planning, and Strategy Reasoning. 

\noindent \textbf{Memory Enhancement.} Revisits and synthesizes the dialogue history to construct a structured summary of the client's state, including their background, presenting issues, and emotional states. This process ensures a continually updated understanding of the client.

\noindent \textbf{Global Planning.} Assesses progress toward therapeutic goals. Plans subsequent steps to ensure that the conversation maintains a coherent therapeutic trajectory, aligning short-term interactions with long-term objectives.

\noindent \textbf{Strategy Reasoning.} Utilizes the outputs from memory enhancement and global planning to determine the optimal response strategy for the current turn. Select specific therapeutic techniques and conversational tactics to address the client's immediate needs while advancing the global plan.

\subsubsection{ MDP Chain-of-Thought Synthesis}
\label{Pattern_Synthesis}

% origin
% To explicitly enhance the decision-making motivations for each turn, we design an iterative optimization method with multi-agent collaboration to synthesize MDP CoT, as shown in Figure~\ref{fig:mdp_pipeline}. Specifically, we utilize the conversation history and key therapy elements as auxiliary information, and sequentially perform memory enhancement, global planning, and strategic reasoning operations through three generative agents, namely, Memory Capture Agent, Global Planning Agent, and Strategy Reasoning Agent, respectively. In addition, the output of each operation is equipped with the same corresponding Checking Agent to verify and correct the generated results to ensure the accuracy and consistency of the generated content. Ultimately, the Fusion Agent integrates the outputs from three agents into a cohesive and comprehensive thought process for the counselor.

To explicitly model the decision-making rationale for each turn, we designed a collaborative multiagent framework to synthesize MDP CoT, as illustrated in Figure~\ref{fig:mdp_pipeline}. This framework sequentially performs memory enhancement, global planning, and strategy reasoning using three specialized generative agents: the Memory Capture Agent, the Global Planning Agent, and the Strategy Reasoning Agent. To ensure accuracy and consistency, the output of each generative step is verified and refined by a corresponding Checking Agent. Finally, a Fusion Agent integrates the validated outputs from the three generative agents into a cohesive and comprehensive thought process. Detailed prompts of different agents are provided in the Appendix~\ref{MDP Prompts}

\noindent \textbf{Memory Capture Agent.}
The task of the Memory Extract Agent is to extract and summarize key memory information based on historical dialogues before the current turn, including (1) an overall description of the client profiles and (2) an assessment of the counseling goal.

\noindent \textbf{Global Plan Agent.}
The Stage Plan Agent combines the memory information examined with the SST knowledge. The agent perform the following two tasks: (1) Identify and confirm the counseling stage that the current dialogue is progressing through; (2) Determine the stage planning required for the current response, which includes three planning states: maintaining the status quo within the current stage, transitioning between steps within the current stage, and facilitating transitions to the next stage. Ultimately, the agent generates a comprehensive global plan.

\noindent \textbf{Strategy Reasoning Agent.}
Combining the content provided by the Memory Capture Agent and the Global Plan Agent, along with the theoretical foundation of SST therapy, the specific response strategy is further refined. The agent selects appropriate expression strategies, such as tone and length, for the current response.

% origin 比较冗余，这里做了精简表述，重新生成的机制需要解释一下？
% The Checking Agent operates across all three stages. In the memory enhancement stage, the Checking Agent assesses whether the counseling goal has been clarified based on the stage labels and compares the accuracy of the analysis regarding the client's information and goal status in the generated content. In the global plan stage, the Checking Agent verifies whether the judgment of historical stages aligns with the stage status label from the previous round of responses and whether the stage planning for the current round is consistent with the current sample's stage label. In the strategy reasoning stage, the Checking Agent analyzes whether the specific policy selected aligns with the stage planning in the global planning and evaluates whether the policy analysis reasonably explains the current round's reply content against the actual response. If any contradictions or deviations between the generated content and the guiding information are found, the Checking Agent will activate a regeneration mechanism to ensure that the content aligns with the expected therapeutic logic and goals. Detailed prompt in Appendix \ref{goal_check} and \ref{strategy_check}.
\noindent \textbf{Checking Agent.}
This agent verifies the output of each generative step to ensure consistency and therapeutic alignment. Specifically, it checks: (1) \textit{Memory Enhancement} for accurate client information and goal status against stage labels; (2) \textit{Global Planning} for logical consistency with historical and current stage labels; and (3) \textit{Strategy Reasoning} for alignment with the global plan and the actual counselor response. If any contradiction is detected, it triggers a regeneration mechanism to ensure that the content aligns with the expected therapeutic logic and goals. Specifically, during the regeneration process, the agent incrementally increases the temperature parameter and incorporates a verification label as part of the prompt—effectively guiding the model toward acceptable outputs. This iterative refinement continues until the generated response passes the validation check or the maximum number of attempts is reached. 

\noindent \textbf{Fusion Agent.}
The Fusion Agent is responsible for integrating the content generated by the first three agents and verifying it with the Checking Agent. The agent then transforms this content into a natural and fluent dialogue in the first person, aligning it more closely with the counselor's actual thought process. An illustrative example of the generated MDP CoT is provided in the Appendix Figure \ref{appendix:mdp_cot_sample}.

\subsection{Conversation LLMs}
\label{develop_LLMs}

% origin 这里的表述被喷说描述得太浮夸，这里修改得朴素一些。以及对condensed SST knowledge的表述做了修改，统一为 the SST knowledge，与后续实验中的表述保持一致。
% Based on the synthesized dataset and thought chains, the multi-turn AI counseling dialogue model CATCH-LLMs with the therapy thought pattern is trained. To enhance the model's understanding of the content, we place the condensed SST knowledge as a system prompt at the beginning of the dialogue sequence.

% In this work, we adopt a standard next-token prediction loss \cite{xie-etal-2025-psydt} to develop CATCH-LLMs. Given an input sequence consisting of the SST knowledge prompt \(\text{K}_{SST}\) and the preceding dialogue history \(\text{K}_{SST}\), the model aims to generate a combined target sequence containing CoT and the counselor’s final response.

% Formally, the loss function is defined as:
% \begin{equation}
% \mathcal{L}_{\text{SFT}} = - \sum_{t=1}^{T} \log P_\theta(y_t \mid y_{<t},\, x),
% \end{equation}
 
% \noindent where \(
% x = concat(\text{K}_{SST}, \text{h})\), \(y = concat(\text{C}_{MDP}, \text{R})
% \), \(\text{R}\) represents counselor’s response, \(\text{C}_{MDP}\) is the MDP CoT. This objective encourages the model to internalize therapy knowledge, generate interpretable thought processes, and produce fluent, contextually aligned responses.

Based on the synthesized dataset and the MDP-CoT, the multi-turn AI counseling dialogue model CATCH-LLMs with the therapy thought pattern is trained. To enhance the model's comprehension of counseling concepts, we place the SST knowledge as a system prompt at the beginning of each dialogue sequence.

In this work, we formulate the training objective of CATCH-LLMs as a standard next-token prediction task \cite{xie-etal-2025-psydt}. Given an input sequence composed of the SST knowledge and the preceding dialogue history, the model predicts a target response that concatenates the MDP-CoT reasoning steps and the counselor’s response.

Formally, the supervised fine-tuning (SFT) loss function is defined as follows:
\begin{equation}
    \mathcal{L}_{\text{SFT}} = - \sum_{t=1}^{T} \log P_{\theta}(y_t \mid y_{<t}, x),
\end{equation}
where:
\begin{align}
    x &= \textit{concat}(K_{\text{SST}}, h), \\
    y &= \textit{concat}(C_{\text{MDP}}, r).
\end{align}

Here, \(P_{\theta}\) represents the probability distribution parameterized by the model parameters \(\theta\), \(T\) denotes the length of the target sequence, \(K_{\text{SST}}\) represents the SST knowledge, \(h\) denotes the dialogue history, \(C_{\text{MDP}}\) represents the MDP CoT, and \(r\) is the counselor’s response.

% 这里的caption是否有必要说明一下指标？
\begin{figure}
    \centering
    \includegraphics[width=3.0 in]{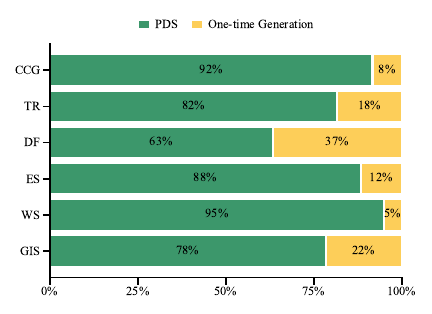}
    \caption{Comparison of expert evaluations between dialogues generated by the PDS strategy and the one-time generation approach. Detailed definitions of each metric are provided in Table \ref{appendix:metrics}.}
    \label{fig:PDS_evaluate}
\end{figure}

\section{Experiments}

\subsection{Implementations}

We employ GPT-4o \footnote{https://chat.openai.com} to implement the PDS strategy for the dialogue dataset, and utilize Doubao-1.5-pro-256k \footnote{https://team.doubao.com} to synthesize MDP CoT, resulting in 233 multi-turn counseling dialogues comprising 6,898 response entries with MDP CoT. Based on the LLaMA-Factory \cite{zheng-etal-2024-llamafactory} framework, we fine-tuned the Qwen3 \cite{qwen3} model at 8B and 14B scales. More training details see in Appendix \ref{appendix:experiments}.

\begin{table}[ht]
\centering
\resizebox{0.4\textwidth}{!}{%
\begin{tabular}{lccc}
        \toprule
 
         & $r$ & $\rho$ & $\tau$ \\
        \midrule
        Correlation & 0.66 & 0.74 & 0.59 \\
        P-value    & 0.0016 & 0.0002 & 0.0009 \\
        \bottomrule
    \end{tabular}
}
  \caption{Correlation coefficients and corresponding p-values between GPT-4o and expert ratings. Here $r$, $\rho$, and $\tau$ denote Pearson’s, Spearman’s, and Kendall’s Tau correlation coefficients, respectively.}
  \label{tab:The correlations between the expert and gpt-4o}
\end{table}

\begin{table*}[ht]
\centering
\resizebox{\textwidth}{!}{%
 \begin{tabular}{lccccccccc}
        \toprule
        \multirow{2}{*}{Model} & \multicolumn{3}{c}{SST-specific Skills} & \multicolumn{3}{c}{General Counseling Skills} \\
        \cmidrule(lr){2-4} \cmidrule(lr){5-7}
        & Solution Focus \(\uparrow \) & Resource Activation \(\uparrow \) & Goal Orientation \(\uparrow \) & Understanding \(\uparrow \) & Interpersonal \(\uparrow \) & Collaboration \(\uparrow \) \\
        \midrule
Qwen3-8B        & 4.47 & 3.98	& 2.85  & 4.47  & \textbf{5.43}  & 4.48 \\
Qwen3-14B       & 4.62	& 4.08	& 3.02  & 4.30   & 5.19	& 4.34 \\
DeepSeek-V3     & 4.83	& 4.78	& 4.04	& 4.53  & 5.04	& 4.94 \\
DeepSeek-R1     & \textbf{4.96}	& 4.79	& 4.07	& 4.69  & 5.31	& 4.91 \\
GPT-4o          & 4.88	& 4.65	& 3.96	& \textbf{4.80}	& 5.39	& 4.96 \\
\midrule
CATCH-8B     & 4.91	&5.29	&4.30 	&4.52	&5.10 	&4.95\\
CATCH-14B    &\textbf{4.96}	&\textbf{5.36}	&\textbf{4.38}	&4.62	&5.17	&\textbf{4.99} \\
        \bottomrule
    \end{tabular}
}
  \caption{Results of evaluation on SST-specific skills and general counseling skills. Scores are averaged over 450 dialogues. The best results are highlighted in \textbf{bold}.}
  \label{tab:main results}
\end{table*}

\subsection{Evaluation Scale}
\label{scale}

\noindent \textbf{SST Therapy Rating Scale.} To evaluate the model’s performance in applying SST principles within counseling dialogues, we adopt the Solution-Focused Inventory (SFI) \cite{grant2012development} as our assessment framework. SST emphasizes the efficient facilitation of clients in a single session to identify and focus on key issues, activate internal and external resources, and formulate clear and achievable goals. These objectives align closely with the three core dimensions of the SFI. Based on this alignment, we define the following SST-specific evaluation metrics derived from SFI: \textbf{Problem Focus (PF)}: Measures the counselor’s ability to guide the client from problem narration toward solution-building, hope, and positive experiences. \textbf{Resource Activation (RA)}: Assesses the counselor’s effectiveness in activating the client’s internal and external resources and resilience, and in translating them into actionable strategies. \textbf{Goal Orientatio (GO)}: Evaluates the counselor’s ability to help the client set clear goals, track progress, and develop feasible action plans. Detailed prompt in Appendix \ref{Evaluation Prompts}.

\noindent \textbf{General Counseling Skill Metrics.} Same as COUNSELINGEVAL \cite{lee-etal-2024-cactus}, we selected three metrics from the CTRS \cite{aarons2012adaptation} to evaluate general counseling skills: Understanding, Interpersonal Effectiveness, and Collaboration.

% 增加补充说明抽取了多少样本，以及具体的评判细则。
\noindent \textbf{Expert vs GPT-4o.} To validate the reliability of LLMs in SFI assessment, we conducted a comparative analysis between expert ratings and GPT-4o's evaluations across three metrics in SST practice. We randomly selected 60 dialogues generated by different models interacting with simulated clients. Experts were recruited to independently rate the transcripts using the same standardized assessment rubric that was provided to GPT-4o in its evaluation prompt. This ensured methodological parity between human and model-based judgments.As shown in Table~\ref{tab:The correlations between the expert and gpt-4o}, GPT-4o demonstrated statistically significant positive correlations with human experts on all three metrics: Pearson correlation coefficient ($r=0.66$, $p=0.0016$), Spearman's rank correlation coefficient ($\rho=0.74$, $p=0.0002$), and Kendall's tau-b coefficient ($\tau=0.59$, $p=0.0009$). These results align with prior findings from Cactus \cite{lee-etal-2024-cactus}, confirming that advanced LLMs like GPT-4o can effectively replicate human-level judgment in assessing key therapeutic competencies required for single-session counseling.

\subsection{Evaluation of PDS Strategy}

To evaluate the effectiveness of the PDS strategy, we employed the same model to generate multi-turn counseling dialogues for 60 randomly sampled counseling cases under the guidance of the SST knowledge, using both the PDS approach and the one-time generation approach. Expert evaluators were then invited to assess and compare the dialogues generated by the two methods for each case. The evaluation was conducted across multiple dimensions, including \textbf{therapeutic implementation effectiveness} (covering the goal-identification stage (GIS), working stage (WS), ending stage (ES)), \textbf{the coherence of the counselor’s guidance} (CCG), \textbf{dialogue fluency} (DF), and \textbf{topic relevance} (TR). For each case, experts were presented with two anonymized dialogues—one generated using the one-time generation approach and the other using the PDS method. Evaluators were instructed to conduct a side-by-side comparison and select the dialogue that demonstrated superior performance on each predefined dimension. Detailed definitions and evaluation criteria for each metric are provided in Appendix \ref{Terminology_explanation}.

\noindent \textbf{Evaluation Results.} As shown in Figure~\ref{fig:PDS_evaluate}, dialogues generated by the PDS strategy demonstrated significantly superior therapeutic implementation effectiveness compared to the one-time generation methods, indicating a higher degree of alignment with therapeutic goals and laying a solid foundation for subsequent MDP synthesis. Moreover, the PDS strategy also obtains outstanding performance in terms of dialogue fluency and topic relevance, suggesting that the PDS strategy contributes to improved logical coherence and semantic consistency, thereby better reflecting the characteristics of real-world counseling interactions.

% 我们的数据集与其他数据集的比较：特定疗法、带有思维链、样本的轮次、语言

\begin{table*}
\centering
\resizebox{\textwidth}{!}{%
\begin{tabular}{@{}llcccccc@{}}
        \toprule
        \multirow{2}{*}{Model} & \multirow{2}{*}{Method} & \multicolumn{3}{c}{SST-specific Skills} & \multicolumn{3}{c}{General Counseling Skills} \\ \cmidrule(lr){3-5} \cmidrule(lr){6-8}
        & & Solution Focus \(\uparrow \) &  Resource Activation \(\uparrow \) & Goal Orientation \(\uparrow \) & Understanding \(\uparrow \) & Interpersonal \(\uparrow \) & Collaboration \(\uparrow \)  \\ \midrule
        \multirow{3}{*}{Qwen3-8B} 
        & w/o PDS and MDP      & 4.47	& 3.98	& 2.85	& 4.47	& \textbf{5.43}	& 4.48 \\
        & w/o MDP  & 4.77	& 5.09	& 4.16	& 4.41	& 4.83	& 4.93 \\
        & CATCH-8B   & \textbf{4.91}	& \textbf{5.29}	& \textbf{4.30} 	
        & \textbf{4.52}	& 5.10 	& \textbf{4.95} \\ 
        
        \midrule
        \multirow{3}{*}{Qwen3-14B} 
        & w/o PDS and MDP     & 4.62	& 4.08	& 3.02	& 4.30	& \textbf{5.19}	& 4.34 \\
        & w/o MDP & 4.81	& 5.14	& 4.14	& 4.45	& 4.91	& 4.92 \\
        & CATCH-14B  & \textbf{4.96}	& \textbf{5.36}	& \textbf{4.38}	
        & \textbf{4.62}	& 5.17	& \textbf{4.99} \\ 
        
        \bottomrule
    \end{tabular}
}
  \caption{Results of the ablation study on SST-specific skills and general counseling skills for Qwen3-8B and Qwen3-14B under various settings. Scores are averaged over 450 dialogues. The best results are highlighted in \textbf{bold}.}
  \label{tab:ablation}
\end{table*}

\subsection{Evaluation for AI Counselor}
\label{evaluation_for_ai_counselor}

\noindent \textbf{Experiment Setup.} We adopt a dialogue simulation evaluation protocol consistent with COUNSELINGEVAL \cite{lee-etal-2024-cactus}, utilizing its accompanying counseling case dataset, which includes 150 client information sheets and three predefined client attitude types (Positive, Neutral, Negative), to assess the application of various models within the SST. Given that the client role primarily involves understanding and responding—tasks that are relatively simple in terms of language generation—we employ GPT-4o-mini as the AI client due to its fast response time and strong language comprehension capabilities, ensuring natural and coherent dialogue generation. We compare five LLMs: GPT-4o, DeepSeek-V3 \cite{deepseekai2024deepseekv3technicalreport} and DeepSeek-R1 \cite{guo2025deepseek}, as well as our backbone models Qwen3-8B \& 14B. To ensure a fair comparison, all method prompts contain the same SST knowledge. Detailed prompts in Appendix \ref{Comparison Prompts}

\begin{figure}
  \centering
  \includegraphics[width=3.0 in]{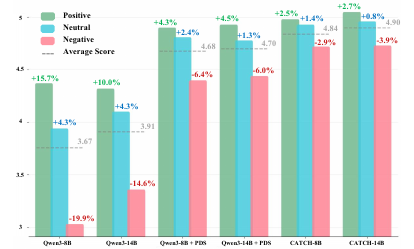}
  \caption{Performance comparison across different client attitudes. The bar chart shows the average evaluation scores of various models under positive, neutral, and negative client attitude settings. The dotted line indicates the overall average score across all attitudes. Notably, baseline models exhibit a pronounced performance gap, particularly struggling with negative-attitude clients. This highlights a critical weakness in real-world counseling scenarios.}
  \label{fig:Attitudes_analysis}
\end{figure}

\noindent \textbf{Main Results.} As shown in Table~\ref{tab:main results}, our model outperforms all baseline models. In particular, our model achieves improvements of \textbf{7.36\%, 31.37\%} and \textbf{45.03\%} over its base models for the SF, RA, and GO metrics, respectively. These results demonstrate that our dataset effectively enhances the model’s ability to apply therapeutic strategies. The improvements in the RA and GO metrics are significantly greater than those in the SF metric. The results suggest that while the base model—when guided by SST knowledge—is already capable of maintaining focus on the client’s problems, it struggles with identifying client resources and sustaining clear goal orientation throughout the dialogue. In terms of general-purpose metrics, our model also maintains strong competitiveness compared to baseline models.

\subsection{Ablation Study} 

% origin
% To evaluate the effectiveness of the PDS strategy and the MDP CoT, we conduct ablation experiments at 8B and 14B scales. Specifically, we fine-tune the base models using multi-turn counseling dialogue data without MDP supervision, yielding variants without explicit planning guidance. These models are then evaluated using the same protocol described in the Evaluation for AI Counselor section. 

% The results are presented in Table 3. When comparing models with and without MDP CoT, we observe consistent performance drops across all three SST-specific skill as well as the three general counseling skill metrics. This indicates that MDP plays a critical role in guiding the model to better understand the counselor's intended therapeutic trajectory at each turn, thereby enhancing not only its capacity to apply SST principles but also its general counseling competency.

% Notably, even without MDP, the models trained with PDS still outperform their backbone models across all SST dimensions. This demonstrates that PDS alone significantly enhances the model’s alignment with therapeutic goals by promoting turn-level coherence and therapy-specific fidelity in synthetic dialogues.

% new 补充多了一个来访者态度的分数分析，体现了我们方法在处理negative的鲁棒性
To systematically evaluate the contributions of the Progressive Dialogue Synthesis (PDS) strategy and the Memory-driven Dynamic Planning Chain-of-Thought (MDP CoT), we conduct ablation experiments on both Qwen3-8B and Qwen3-14B models. We compare the full CATCH framework against variants trained with partial or no access to our proposed components, using the same evaluation protocol as described in Section \ref{evaluation_for_ai_counselor}. The results are summarized in Table~\ref{tab:ablation}.

\noindent \textbf{Impact of MDP CoT.} When comparing models with and without MDP CoT, we observe consistent performance drops across all three SST-specific skill as well as the three general counseling skill metrics. This indicates that MDP plays a critical role in guiding the model to better understand the counselor's intended therapeutic trajectory at each turn, thereby enhancing not only its capacity to apply SST principles but also its general counseling competency.

\noindent \textbf{Contribution of PDS.} Notably, even without MDP, the models trained with PDS still outperform their backbone models across all SST dimensions. This demonstrates that PDS alone significantly enhances the model’s alignment with therapeutic goals by promoting turn-level coherence and therapy-specific fidelity in synthetic dialogues.

\noindent \textbf{Robustness Across Client Attitudes.} We further analyze model performance under different client attitudes—positive, neutral, and negative—to assess robustness in handling challenging interactions. As shown in Figure~\ref{fig:Attitudes_analysis}, baseline models (w/o PDS and MDP) exhibit a pronounced performance gap, particularly struggling with negative-attitude clients. This highlights their vulnerability to therapy drift when confronted with complex or non-compliant client dynamics.

% \noindent \textbf{Attitudes Analysis.} We also analyze the model's performance across different client attitudes (positive, neutral, negative) to evaluate its robustness in handling challenging counseling scenarios. As shown in Figure \ref{fig:Attitudes_analysis}, the baseline models (Qwen3-8B/14B w/o PDS and MDP) exhibit a significant performance gap, particularly struggling with clients exhibiting a negative attitude. This suggests that without explicit planning guidance and structured dialogue synthesis, models are highly susceptible to therapy drift when faced with complex client dynamics. In contrast, our CATCH framework demonstrates superior robustness. Both the integration of PDS and the MDP CoT contribute to substantial performance gains across all attitudes, with the most pronounced improvements observed in the negative attitude setting. This indicates that the structured reasoning and stage-aligned generation enabled by CATCH provide a crucial cognitive scaffold, allowing the model to maintain therapeutic fidelity and effectively navigate difficult interactions, thereby delivering more consistent and reliable counseling outcomes.

\section{Conclusion}
\label{sec:bibtex}

In this work, we introduced a novel data synthesis framework, CATCH, designed to enhance the quality of multi-turn counseling dialogues in AI counseling. By employing the PDS strategy, we generated a counseling dialogue dataset that aligns with therapeutic intervention principles, thereby improving therapy fidelity. Additionally, the MDP CoT clarified decision-making motivations for each dialogue turn, facilitating a deeper understanding of therapeutic logic. Our extensive experiments demonstrate that CATCH not only enhances the fidelity and logical coherence of AI counseling but also effectively mitigates therapy drift. This framework represents a significant advancement in the field of AI counseling, providing a robust foundation for future research and applications. 

\section*{Limitations}

% 写得精简一些了
While our results affirm the CATCH framework's effectiveness in enhancing counseling models, several limitations warrant consideration. One key consideration is the scale of our training corpus. Our collection of 233 multi-turn dialogues, though meticulously aligned with SST principles, may not be extensive enough to ensure robust generalization across a wide spectrum of psychosocial contexts and linguistic styles.

Moreover, the dataset is confined to single-session interactions, thereby neglecting the longitudinal dynamics inherent in authentic therapeutic relationships. The current architecture lacks a mechanism for cross-session memory integration, which is critical for tracking patient progress, adjusting interventions, and managing the evolution of therapeutic goals. Future research will therefore prioritize both the expansion of our dataset and the development of hierarchical memory architectures to enable coherent, multi-session therapeutic tracking.

\section*{Acknowledgments}

This work was supported by the Guangdong Basic and Applied Basic Research Foundation (2025A1515011203), Guangdong Provincial Key Laboratory of Human Digital Twin (2022B1212010004), and Nansha Key Project under Grant (2022ZD011).

\bibliography{custom}

\newpage

\appendix

\section{Terminology and Assessment Metrics}
\label{Terminology_explanation}

\noindent\textbf{Terminology in Single-Session Therapy (SST).} We provide detailed explanations of key terminology and definitions of evaluation metrics to ensure a consistent understanding of core concepts and assessment criteria. Table \ref{appendix:terms_part1} and \ref{appendix:terms_part2} presents clear definitions of essential psychotherapeutic terms used in SST. These terms constitute the foundational framework of SST and inform decision-making throughout the data synthesis process. 

\noindent\textbf{Metrics and Criteria in Evaluation of PDS.} Table \ref{appendix:metrics} elaborates on the six core dimensions employed in PDS evaluation: Goal Identification Stage Performance (GIS), Working Stage Performance (WS), Ending Stage Performance (ES), Dialogue Fluency (DF), Topic Relevance (TR), and Counselor’s Coherent Guidance (CCG). Each dimension is accompanied by specific evaluation criteria and rating guidelines to ensure the reliability and validity of the assessment outcomes. 

% 术语的解释表格1
\begin{table*}{
\centering
\renewcommand\arraystretch{1.3}
\begin{tabular}{|p{2.0cm}|p{12cm}|}
\hline
\textbf{Terms}  & \textbf{Description} \\
\hline
Counseling Goal & In Single-Session Therapy (SST), the counseling goal refers to a specific, mutually agreed-upon direction of psychological change that the client and therapist aim to achieve within one session. It must be clear, actionable, and measurable, focusing on the client’s internal psychological processes—such as emotional regulation, cognitive patterns, behavioral habits, or interpersonal dynamics—rather than on solving external real-world problems (e.g., financial stress, work conflicts). The goal reflects a shift in how the client experiences and responds to challenges, not merely the resolution of the challenges themselves. By targeting underlying psychological mechanisms, the counseling goal enables meaningful, client-centered change while ensuring therapeutic focus and coherence.  \\
\hline
Resources & In SST, resources refer to the conditions and capabilities available to the client—either within themselves or in their surrounding environment—that support their ability to cope with difficulties, promote psychological growth, and achieve the counseling goal. These are categorized into internal and external resources. Internal resources include emotional regulation abilities, past successful coping experiences, personal strengths, and resilience. External resources encompass social support systems (such as family, friends, colleagues, and community groups), professional services, and environmental factors conducive to problem-solving. Positive resources are those that can facilitate goal attainment, while non-positive resources may involve ineffective coping mechanisms or unsupportive environments, which need to be identified and modified accordingly. \\
\hline
Solution & The solution in an SST session is a concrete and feasible action plan that emerges from the activation and integration of the client’s positive resources in response to the counseling goal. Rather than being generic advice, the solution is developed through therapist guidance and client participation, including rehearsal and practical application, to ensure it is both actionable and empowering in real-life contexts. Solutions emphasize effectiveness and client autonomy, enabling the client to benefit from the intervention without reliance on long-term therapy, thus fostering effective problem alleviation or resolution. \\
\hline
Goal Confirmation Stag & The goal confirmation stage marks the beginning of SST and involves establishing a strong therapeutic alliance and collaboratively identifying a clear and achievable counseling goal. This stage adheres to the “less is more” principle, focusing on the most urgent and representative core issue. Through empathy, encouragement, and inquiry, the therapist assists the client in articulating their needs and, based on a thorough understanding of the client’s background and emotional state, negotiates and agrees upon a goal. The goal must be specific, constructive, and assessable, grounded in psychological perspectives, thereby setting a solid foundation for the work that follows. \\
\hline
\end{tabular}
\caption{Table of Terminology Explanations}
\label{appendix:terms_part1}
}
\end{table*}

% 术语的解释表格2
\begin{table*}{
\centering
\renewcommand\arraystretch{1.3}
\begin{tabular}{|p{2.0cm}|p{12cm}|}
\hline
\textbf{Terms}  & \textbf{Description} \\
\hline
Working Stage & The working stage is the core phase of SST and comprises two primary steps: exploring resources and implementing actions. During this stage, the therapist helps the client identify their internal and external positive resources and convert them into concrete solutions through techniques such as role-play and behavioral rehearsal. Emphasis is placed on client engagement and initiative, with the therapist serving as a guide and supporter. The aim is to help the client uncover their potential and build confidence in their problem-solving abilities. When resources are effectively activated and the client demonstrates a willingness to change and to try specific actions, the process can proceed to the final stage. \\
\hline
Ending Stage & The ending stage concludes the SST process and occurs once the counseling goal has been largely achieved or a significant breakthrough has been made. The primary tasks in this phase include summarizing and providing feedback on the session, reinforcing the client’s positive experiences and accomplishments, and assigning follow-up tasks to promote the application and consolidation of gains in real-life settings. Depending on the client’s needs, the therapist may offer recommendations for ongoing support or further counseling. This stage emphasizes completeness and constructiveness, aiming to ensure that the client leaves with a sense of efficacy and readiness to face future challenges independently. \\
\hline
Treatment Fidelity & Treatment fidelity refers to the extent to which a psychological intervention is delivered in accordance with its theoretical model, manualized protocol, and core therapeutic components. It ensures that key procedures—such as goal setting, stage progression, and technique application—are implemented consistently, accurately, and as intended, minimizing drift or deviations during practice. High treatment fidelity is critical for maintaining therapeutic integrity, enabling reliable replication, and ensuring that observed outcomes are attributable to the intervention itself rather than ad hoc or inconsistent delivery. \\
\hline
\end{tabular}
\caption{Table of Terminology Explanations}
\label{appendix:terms_part2}
}
\end{table*}

\begin{table*}
  \centering
  \renewcommand\arraystretch{1.3}
  \begin{tabular}{|p{3.0cm}|p{4.0cm}|p{7.8cm}|}
  \hline
  \textbf{Indicator} & \textbf{Definition} & \textbf{Evaluation Criteria} \\
  \hline
  Goal Identification Stage Performance (GIS) & Examines whether the process of collaboratively identifying the counseling goal in the initial stage meets SST requirements. & Whether the counselor can establish a good relationship through effective communication, using empathy, genuineness, and unconditional positive regard to build basic trust; whether the counselor follows the "less is more" principle to help the client identify the most urgent and key issue; whether the final goal meets the criteria of being specific, feasible, positive, mutually agreed, measurable, and within the scope of psychology rather than a purely practical issue. \\
  \hline
  Working Stage Performance (WS) & Evaluates the counselor’s performance during the resource-searching and action-rehearsal phase. & Whether the counselor explores internal resources before external ones (or alternately), sufficiently investigates the client's internal strengths (e.g., past coping efforts, resilience), encourages effective efforts and modifies ineffective ones; whether the counselor explores external support by identifying people who understand the client’s difficulties and their advice, and helps transform external support into usable resources; whether action rehearsal is smoothly initiated when readiness is shown, and a workable solution is practiced or assigned for after-session use. \\
  \hline
  Ending Stage Performance (ES) & Assesses the appropriateness of summary, feedback, encouragement, and follow-up suggestions in the final stage. & Whether the counselor gives appropriate feedback and encouragement (e.g., praising motivation and previous efforts), accurately summarizes the session, assigns targeted homework, encourages real-life actions, and when necessary, suggests continued counseling or external help, ensuring the session feels complete and constructive. \\
  \hline
  Dialogue Fluency (DF) & Refers to the coherence and smoothness of the dialogue. & Dialogue should transition naturally from one topic to another, counselor’s questions and responses should be relevant to the client’s statements, and sentences should be clear and fluent without noticeable pauses, repetitions, contradictions, or illogical jumps. \\
  \hline
  Topic Relevance (TR) & Examines whether the dialogue stays focused on the client’s issues or counseling goals. & All content, including questions, suggestions, and feedback, should closely relate to the client’s core concerns without digressions, maintaining a high level of focus and relevance. \\
  \hline
  Counselor’s Coherent Guidance (CCG) & Refers to the logical consistency and clarity of the counselor’s guiding process throughout the session. & The counselor should guide the client in a logically ordered manner—from relationship-building and goal-setting to resource exploration and action rehearsal, ending with summary and feedback. Each phase should connect naturally, helping the client think progressively and move toward resolution. \\
  \hline
  \end{tabular}
  \caption{Evaluation metrics for the PDS strategy and the one-time generated dialogue.}
  \label{appendix:metrics}
  \end{table*}

% 对话样本1
\begin{figure*}
    \centering
    \includegraphics[width=0.8\linewidth]{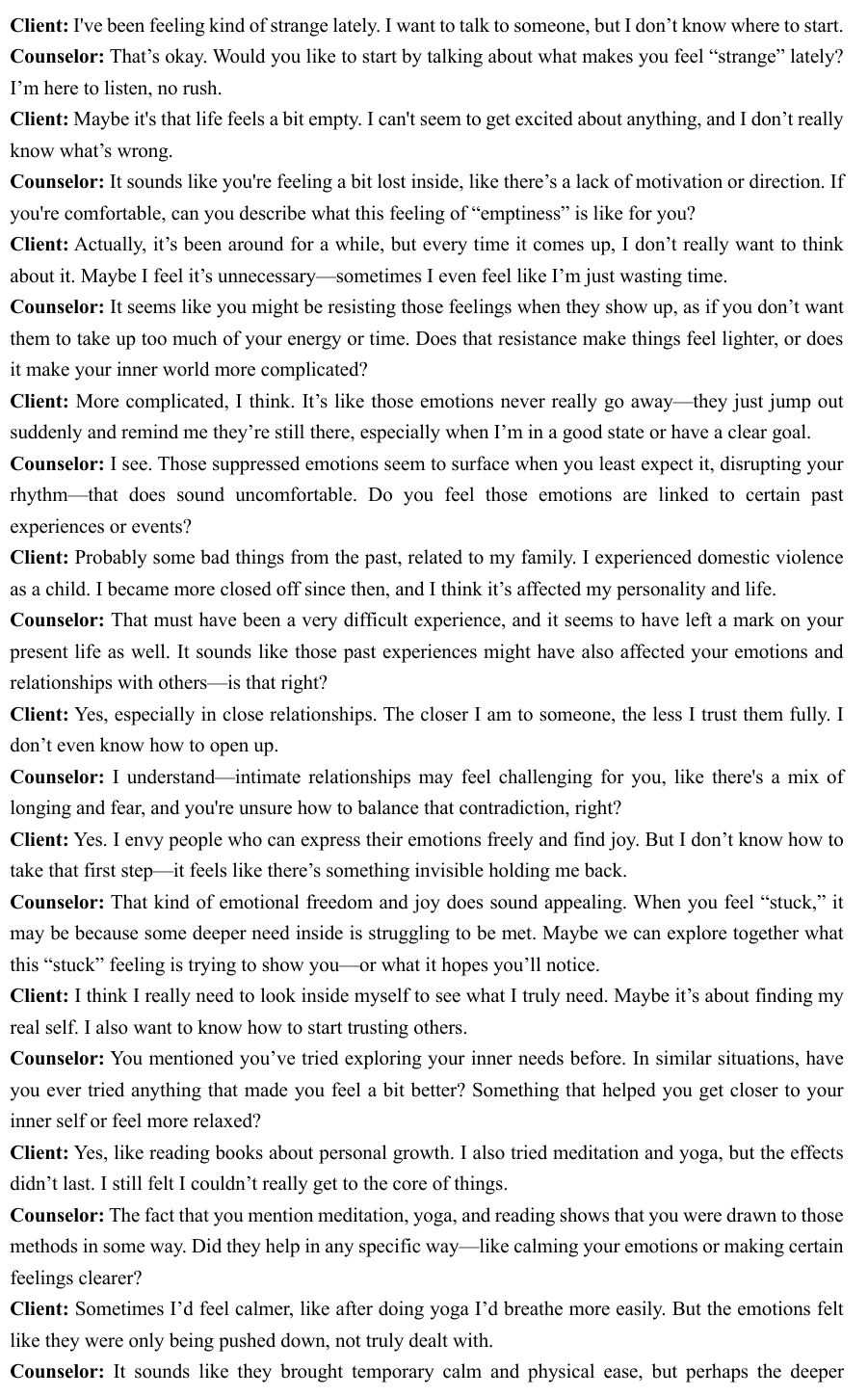}
    \caption{PDS dialogue sample page 1 of 3.}
    \label{Sample_dia1}
\end{figure*}

% 对话样本2
\begin{figure*}
    \centering
    \includegraphics[width=0.8\linewidth]{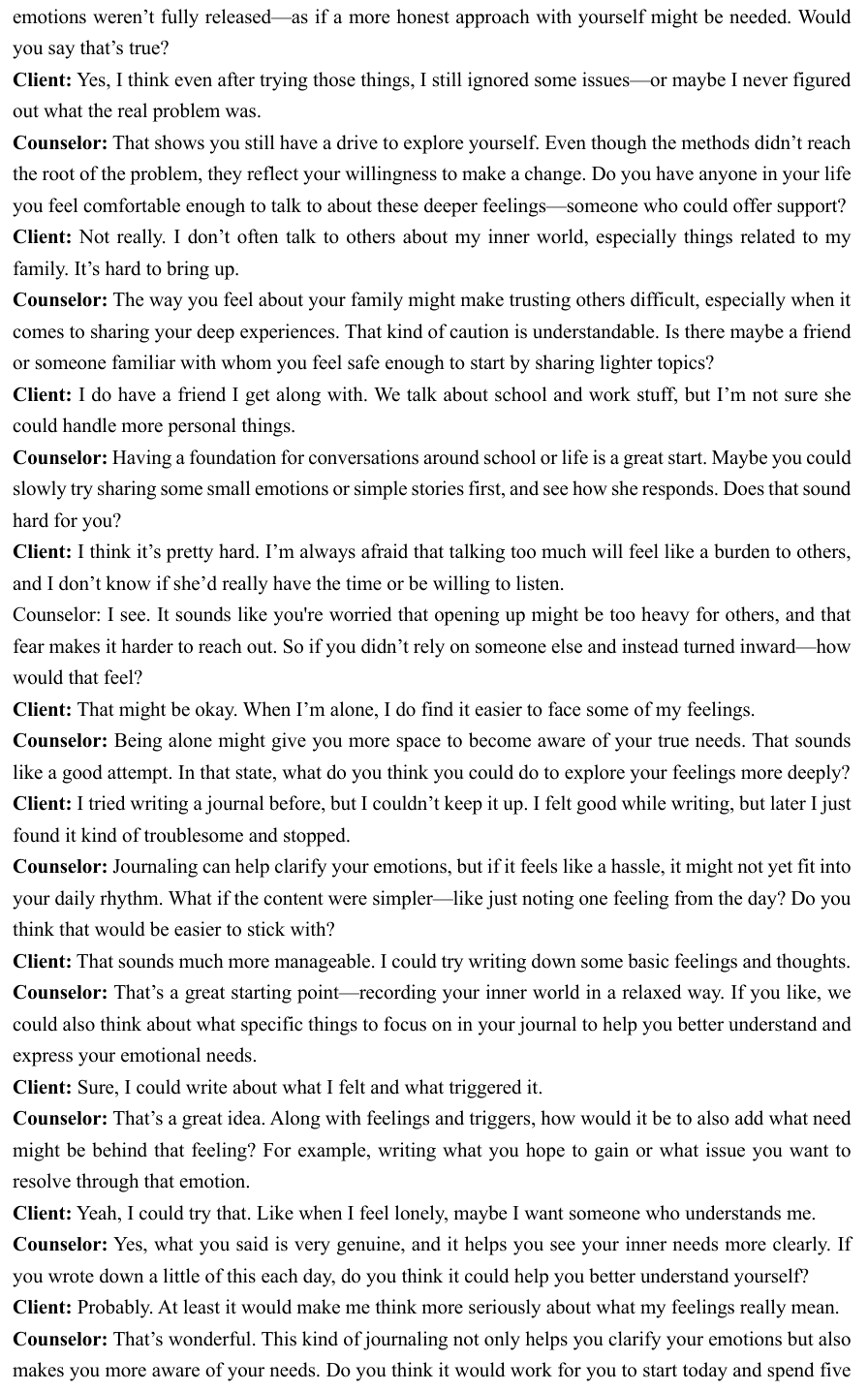}
    \caption{PDS dialogue sample page 2 of 3.}
    \label{Sample_dia2}
\end{figure*}

% 对话样本3
\begin{figure*}
    \centering
    \includegraphics[width=0.8\linewidth]{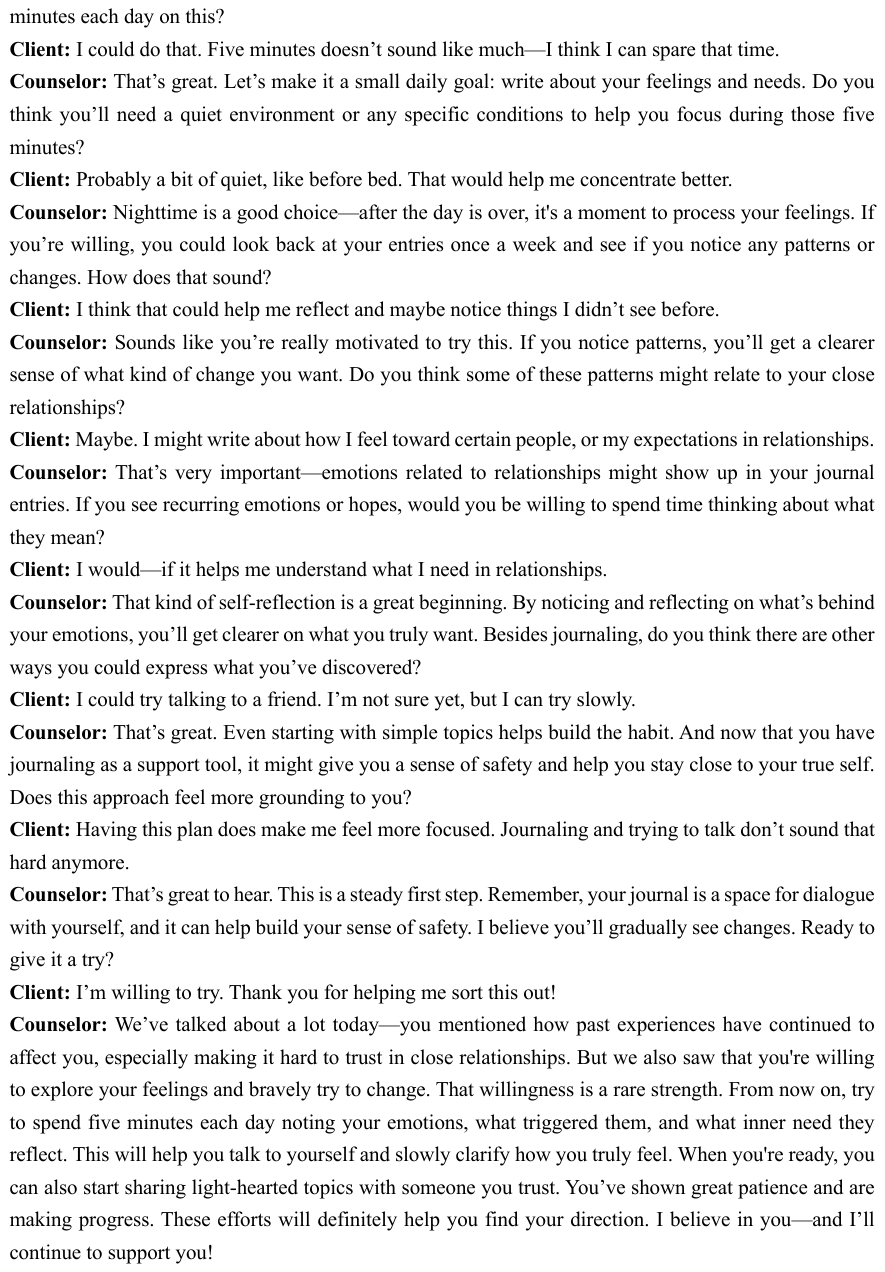}
    \caption{PDS dialogue sample page 3 of 3.}
    \label{Sample_dia3}
\end{figure*}

% MDP样本
\begin{figure*}
    \centering
    \includegraphics[width=1.0\linewidth]{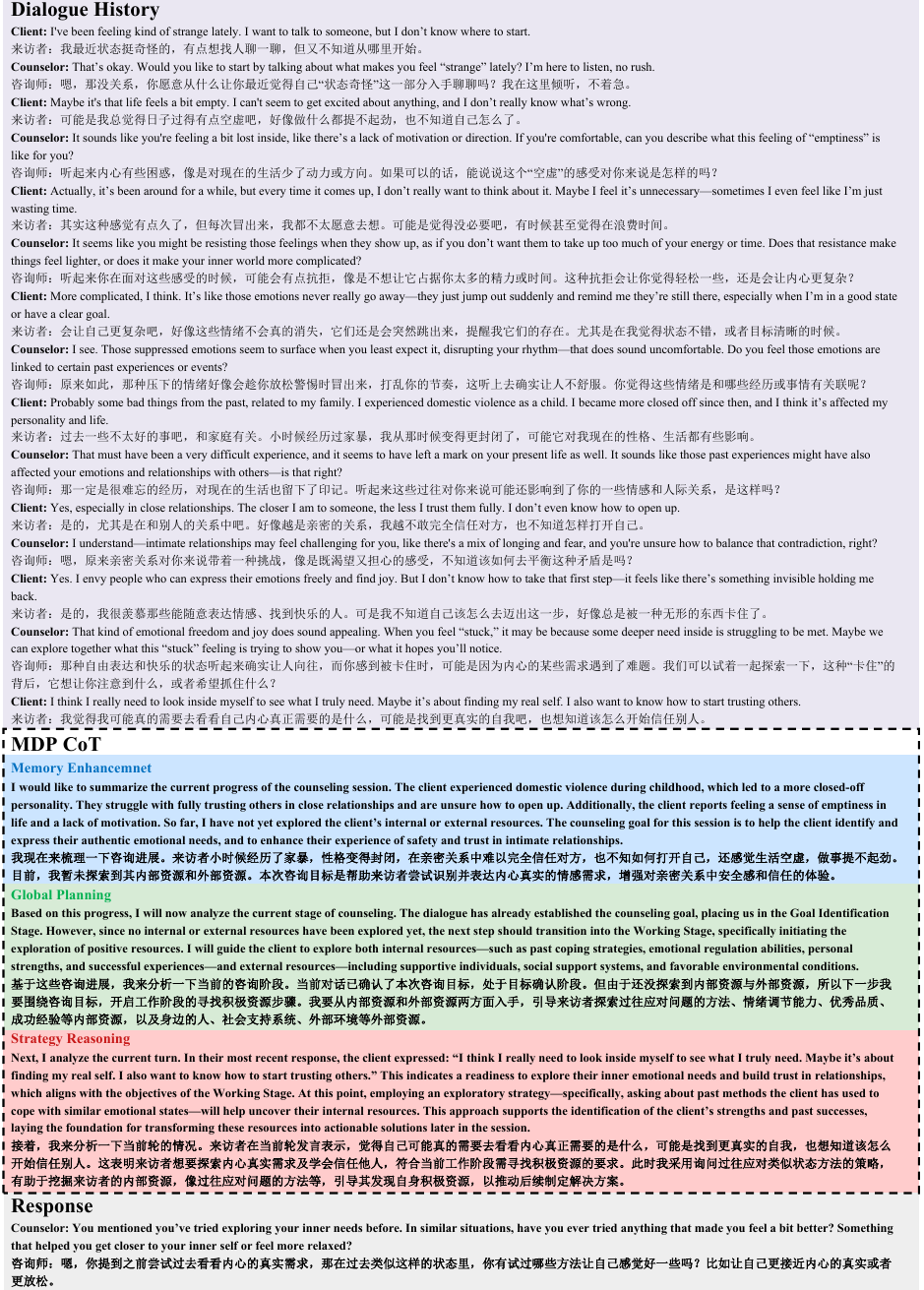}
    \caption{MDP CoT Example.}
    \label{appendix:mdp_cot_sample}
\end{figure*}

\section{Details of Experiments}
\label{appendix:experiments}

\noindent\textbf{Training Details.} We fine-tuned the Qwen3 models using the Low-Rank Adaptation (LoRA)\cite{hu2021loralowrankadaptationlarge} method to efficiently adapt the pre-trained language model to the counseling domain. The optimization was performed using the AdamW optimizer\cite{loshchilov2017decoupled} with a learning rate of $1e-4$ and a cosine learning rate scheduler. We set the warm-up ratio to $0.1$ to stabilize the initial training phase. The LoRA rank was set to $16$ and the global batch size was set to $32$. The models were trained for $3$ epochs on $2$ NVIDIA A800 80GB GPUs using mixed-precision training (FP16) to improve computational efficiency. All training configurations were kept consistent across the 8B and 14B model variants.

\noindent\textbf{Inference Details.} During inference, we adopt a temperature of $0.6$ for all AI counselor models as well as for the AI client simulated by GPT-4o-mini. For the AI evaluator, we set the temperature = $0$. To improve inference efficiency and reduce latency, we leverage the vLLM library \footnote{https://docs.vllm.ai}.

\newpage
\newpage

\onecolumn

\section{Prompts}

\subsection{Prompts for PDS Strategy.}
\label{PDS Prompts}

% The prompt for generating counseling goal in PDS
\begin{tcolorbox}[
colback=mycolor2,
colframe=mycolor,
title=\textbf{The prompt for generating counseling goal in PDS}, 
coltitle=black,
fonttitle=\bfseries\large,
fontupper=\normalsize\rmfamily,
breakable
]
You are an excellent counselor specializing in Single Session Therapy (SST). Based on the client’s self-report, your task is to formulate a goal for this session (summarized in 1–2 sentences). \\

\#\# Client’s Self-report \\
\textbf{[client info]} \\

\#\# Notes \\
1. The goal must meet the following criteria: \\
- **Unique**: Focus on only one core (most urgent and critical) issue to avoid dispersing direction and energy by addressing multiple issues simultaneously. \\
- **Specific**: The goal must be clear and concrete. It should explicitly describe the desired change in behavior, emotion, or cognition. \\
- **Feasible**: The goal should be achievable within the client's existing resources, abilities, and time. It must be grounded in the client’s self-report. \\
- **Positive**: The goal should be expressed in positive language, emphasizing the desired positive state rather than the avoidance or elimination of negative conditions. \\
- **Measurable**: The goal should be able to be evaluated or verified at the end of the session, based on observable behaviors, emotional changes, or cognitive shifts.\\
- **Psychological, not practical**: The goal should focus on the client’s internal psychological domain, such as emotion regulation, cognitive change, or behavioral pattern adjustment, rather than solving external practical problems.\\

2. The formulated goal should be concise and clear, avoiding complex terminology or overly technical language. It should be easy to understand and remember, without explaining how to assess it. \\

\#\# Output format (strictly follow): \\
goal:...
\end{tcolorbox}
  
% The prompt for generating resources and solution in PDS
\begin{tcolorbox}[
colback=mycolor2,
colframe=mycolor,
title={\textbf{The prompt for generating resources and solution in PDS}}, 
coltitle=black,
fonttitle=\bfseries\large,
fontupper=\normalsize\rmfamily,
breakable
]
You are an excellent supervisor in SST (Single Session Therapy). Your task is to create a supplemental counseling case based on the requirements below and return the results in the specified JSON format: \\

\#\# Client's Self-Report \\
\textbf{[client info]} \\

\#\# Client Personality Traits \\
\textbf{[personality info]} \\

\#\# Counseling Goal \\
\textbf{[goal info]} \\

\#\# Resource Explanation \\
1. Definition of Internal and External Resources: \\
  - Internal Resources: The client’s past problem-solving strategies, emotional regulation techniques, personal interests, and other internal strengths.\\
  - External Resources: The client’s social network (family, friends, colleagues, etc.), community or group support systems, and available life/work conditions. \\

2. Definition of Positive and Non-Positive Resources: \\
  - Non-Positive Resources: Resource directions explored during the session but could not be effectively transformed into action plans due to various reasons, such as the client lacking the resource or being resistant to it. \\
  - Positive Resources: Resources that directly support achieving the counseling goal and can be deeply explored and transformed into specific actions. \\

\#\# Task Requirements: Dialogue Element Creation \\
1. Solution Conception: \\
  - First, based on the [Client's Self-Report], [Personality Traits], and [Counseling Goal], devise one solution. \\
  - Ensure the solution is specific, feasible, entirely executable by the client, and positively contributes to the counseling goal.\\
  - Ensure the solution fits the client’s background information and is closely aligned with the goal.
  - Avoid vague or unrealistic suggestions (e.g., “just take deep breaths” or “find another counselor”). \\

2. Resource Planning and Creation: \\
  - Based on the chosen best solution, identify one key **positive resource** that directly supports the solution’s execution. \\
  - Then create **\textbf{[negative num]} non-positive resources**: these are resources explored during counseling but ultimately could not be utilized. Describe how the counselor attempted to explore them, the method of exploration, why they failed to be converted, and how the focus shifted to the next resource. \\
  - Ensure smooth logical transitions between resources, typically starting with internal resources and then moving to external ones. \\
  - Only one positive resource is allowed (as the final item), which should be clearly linked to the action plan. \\

3. Integration of Resources and Solution: \\
  - Ensure there is a coherent flow between resources to facilitate dialogue construction. \\
  - Ensure a clear and reasonable connection between the positive resource and the solution, and describe in detail how the positive resource is transformed into a concrete, feasible action plan. \\
  - Keep the language concise and clear; avoid overly technical or professional jargon. \\

\#\# Output Format \\
- Strictly follow the JSON format below, and fill in each field with your analysis:
\begin{lstlisting}[
    basicstyle=\normalsize\ttfamily,
    breaklines=true,
    frame=none,
    xleftmargin=0.5cm,
    xrightmargin=0.5cm,
    aboveskip=0.5em,
    belowskip=0.5em,
    numbers=none
]
{
  "resource": [
    {
      "label": "Resource label: 1, 2, 3, etc.",
      "type": "internal/external",
      "valence": "positive/negative", 
      "description": "Resource description (10-20 characters)",
      "relevance": "Resource's relevance to the counseling goal (30-50 characters)",
      "next_label": "Next resource label: 2, 3, etc. (null for the last one)",
      "change": "Detailed description of the transition from this resource to the next or how it's turned into an action (>80 characters)"
    }
  ],
  "solution": "{Solution description}"
}
\end{lstlisting}

\end{tcolorbox}

% The prompt for generating dialogue in the goal identification stage in PDS
\begin{tcolorbox}[
colback=mycolor2,
colframe=mycolor,
title=\textbf{The prompt for generating dialogue in the goal identification stage in PDS}, 
coltitle=black,
fonttitle=\bfseries\large,
fontupper=\normalsize\rmfamily,
breakable
]
\#\# Example dialogue and analysis \\
\textbf{[example]} \\

\#\# **Background Information** (Note: The counselor starts with no knowledge of this information and must naturally elicit it during the dialogue) \\
Client's Self-Report \\
\textbf{[client info]} \\
Client Personality Traits \\
\textbf{[personality info]} \\
Counseling Goal \\
\textbf{[goal info]} \\

\#\# **Task** \\
Based on the provided background information of the client, simulate a realistic counseling session following the logic of SST therapy, generating a multi-turn dialogue for the **Goal Clarification Phase** (**\textbf{[turn num]} turns**). \\

\#\#\# **[SST Goal Clarification Phase Process]** \\
- **Core Task**: Build trust and gather information to eventually identify the core goal of this counseling session (unique, specific, and of greatest concern to the client).  \\
- **Key Steps**: \\
  - **Greeting**: The counselor and client exchange simple greetings to create a relaxed atmosphere. \\
  - **Building Rapport**: The counselor initiates the conversation with warmth and empathy, avoiding direct discussion of problems. \\
  - **Information Gathering**: The counselor uses open-ended questions to gradually collect the client’s basic information in preparation for goal clarification. \\
  - **Goal Clarification**: In the final round of this phase, the counselor and client work together to clearly identify the specific goal of this session (unique, specific, and most important to the client). Ensure alignment with the **Counseling Goal** in the background information. \\
- **Prohibited**: The counselor must not directly quote or reference content from the **background information**, nor should they directly state the **Counseling Goal**; it must be naturally revealed through dialogue. \\

\#\#\# **[Language and Dialogue Rules]** \\
1. **Counselor's Language**: \\ 
- Natural, human-like, warm, and conversational, avoiding technical or clinical terms. Refer to the style of the **Example Dialogue and Analysis**.\\
- Avoid closed-ended questions, and primarily use open-ended questions to encourage the client to elaborate. \\
- Each turn should be no more than two sentences, and must be closely tied to the client’s previous statement. \\
- Use empathy, encouragement, and concretization techniques. Avoid invalidating the client’s feelings. \\
2. **Client's Language**: \\
- Should reflect a realistic counseling setting, gradually revealing information in response to the counselor's prompts; do not disclose all background details at once.\\
- Client responses must stay closely connected to the counselor’s remarks and align with their background information.\\
- The [Client Personality Traits] are based on the Big Five personality theory, and the client’s communication style throughout the dialogue should reflect these traits.\\
3. **Number of Dialogue Turns**:\\
- Ensure the dialogue has exactly **\textbf{[turn num]} turns**, with each turn consisting of one statement by the client and one by the counselor.\\
- The client's statement should be both the **starting** and **ending** turn of this phase.\\

\#\# **Output Format Requirements**: Strictly follow the format and target number of turns below \\
Turn 1: \\
Client: ... \\
Counselor: ... \\
Turn 2: \\
... \\
Turn \textbf{[turn num]}: \\
Client: ... \\
\end{tcolorbox}
  
% The prompt for generating dialogue in the working stage in PDS
\begin{tcolorbox}[
  colback=mycolor2,
  colframe=mycolor,
  title=\textbf{The prompt for generating dialogue in the working stage in PDS}, 
  coltitle=black,
  fonttitle=\bfseries\large,
  fontupper=\normalsize\rmfamily,
  breakable
  ]
\#\# Example Dialogue and Analysis\\
\textbf{[example]}\\

\#\# **Background Information** (Note: The counselor is completely unaware of this information at the beginning and must naturally elicit it through the dialogue)\\
Client Personality Traits\\
\textbf{[personality info]}\\
Counseling Goals\\
\textbf{[goal info]}\\
Resource Exploration Path\\
\textbf{[resource]}\\
Solution\\
\textbf{[solution]}\\

\#\# **Task**\\
Simulate a realistic counseling scenario that follows the logic of SST (Solution-Focused Brief Therapy), and continue generating multi-turn counseling dialogue for the **Working Phase** based on the **Goal Clarification Phase Dialogue** and guided by the **Resource Exploration Path** (**\textbf{[turn num]} turns**).\\
- **Note**:\\
  - Follow the **Resource Exploration Path** in strict order (no omissions): start with **non-positive resources**, then explore other resources, and finally explore **positive resources** and transform them into behavioral rehearsals.\\
  - The counselor must naturally elicit information through conversation, rather than referencing the background directly.\\
  - Do not mention any content unrelated to the client's background.\\
  - The counselor must avoid language implying the end of the session (e.g., "goodbye," "talk to you next time," etc.)\\

\#\#\# **Language and Dialogue Rules**\\
1. **Counselor Language**:\\
- Use natural, warm, and conversational expressions.\\
- Refer to the counselor's tone in the **Example Dialogue and Analysis** section, but do not directly copy its language.\\
- Avoid professional jargon (e.g., "resources," "behavioral rehearsal," "solution rehearsal," etc.)\\
- Avoid closed-ended questions; prefer open-ended ones.\\
- Each turn should not exceed two sentences (<50 words), and should closely respond to the client's previous message.\\
- Use techniques like empathy, encouragement, and concretization, and avoid invalidating the client's feelings.\\

2. **Client Language**:\\
- Reflect a realistic counseling setting, revealing information progressively.\\
- Responses should closely relate to the counselor's questions and their own background.\\
- Language style must consistently reflect the **Client Personality Traits**.\\

3. **Dialogue Turns**:\\
- Ensure there are a total of \textbf{[turn num]} turns (each turn consists of one counselor utterance and one client response).\\
- The Working Phase begins with a counselor's utterance and ends with a client's response.\\
- Ensure a natural transition from the last line of the **Goal Clarification Phase Dialogue**.\\

\#\#\# Goal Clarification Phase Dialogue\\
\textbf{dialogue}\\

\#\#\# **Output Format Requirements**: Strictly follow the format and required number of turns below, and generate the dialogue until turn \textbf{[turn num]}.\\
Counselor: ...\\
Turn \textbf{[begin turn]}:\\
Client: ...\\
Counselor: ...\\
...\\
Turn \textbf{[turn num]}:\\
Client: ...\\
\end{tcolorbox}

% The prompt for generating dialogue in the ending stage in PDS
\begin{tcolorbox}[
  colback=mycolor2,
  colframe=mycolor,
  title=\textbf{The prompt for generating dialogue in the ending stage in PDS}, 
  coltitle=black,
  fonttitle=\bfseries\large,
  fontupper=\normalsize\rmfamily,
  breakable
  ]
\#\# **Background Information** \\  
Counseling Goals \\  
\textbf{[goal info]} \\  
Solution \\  
\textbf{[solution]} \\  
Dialogue History \\  
\textbf{[dialogue]} \\  

\#\# **Task** \\  
Based on the **Dialogue History**, simulate a realistic counseling scenario consistent with the logic of SST (Social Skills Training) and generate the counselor’s concluding feedback during the **Ending Phase**: \\  

\#\#\# **SST Ending Phase Summary Statement** \\  
- **Core Task**: Provide a feedback summary and encouragement for the session, affirm the client’s strengths, and assign a homework task to help maintain the progress made. \\  
- **Include the following elements**: \\  
  - **Feedback Summary**: Briefly recap the client’s main struggles and the helpful elements they’ve discovered (e.g., “You mentioned feeling guilty after arguing with your child, but also realized how much you care and that you’ve taken steps in the past to make peace”), and emphasize their **motivation to change** (e.g., “You shared a lot of details today, and I can really see how much you want to improve your relationship with your child—that care itself is such a strong force.”). \\  
  - **Encouragement for Action**: Assign a concrete and achievable homework task (consistent with the **Proposed Solution** from the working phase). Ensure the task fits the client’s actual situation and goals. The language should be natural, warm, and non-technical. \\  
  - **Emotional Connection**: Express understanding and support, affirm the client’s effort and capacity for change, and offer encouragement. \\  

\#\#\# **Important Notes** \\  
- The response must include all three parts: summary, homework, and emotional connection, and should be a **single, long reply (3–5 sentences)**. \\  
- Only generate the **counselor’s response**, do not include any headings or additional content. \\  
- **Avoid all professional jargon** (e.g., “resources,” “rehearsed actions,” “solution rehearsal”); use natural, heartfelt, conversational language instead. \\  

\#\#\# **Output Format Requirement**: Strictly follow the format below \\  
counselor: ... \\  
\end{tcolorbox}

% The prompt for checking dialogues structure in PDS
\begin{tcolorbox}[
  colback=mycolor2,
  colframe=mycolor,
  title=\textbf{The prompt for checking dialogues structure in PDS}, 
  coltitle=black,
  fonttitle=\bfseries\large,
  fontupper=\normalsize\rmfamily,
  breakable
  ]
You are a supervisor of Single-Session Therapy (SST) and are responsible for reviewing the dialogue records of psychological counselors. Based on the **Single-Session Therapy (SST) Guideline Manual**, assess the quality of the counseling dialogue according to the following checklist and return the results in the specified JSON format: \\

\#\# Counseling Goal \\
\textbf{[goal info]} \\

\#\# Counseling Dialogue \\
\textbf{[dialogue]} \\

\#\# Checklist \\
1. goal identification Phase: Did the counselor first establish a good relationship with the client, and by the end of this phase, jointly agree on a goal with the client that aligns with the **Counseling Goal**? \\
2. Working Phase: \\
- Did the counselor **thoroughly explore** the client's internal and external resources instead of prematurely moving into action rehearsal? \\
- Did the counselor conduct action rehearsal when transitioning to the **Ending Phase**? \\
- Did the counselor maintain the dialogue focused on the **Counseling Goal** throughout this phase? \\
3. Ending Phase: Did the counselor provide feedback, encouragement, and affirmation of the client's strengths? Did the counselor assign homework, ensure that the summary content matched the previous dialogue, and that the assigned homework aligned with the action rehearsal? \\

\#\# Output Format Specification \\
- Output must strictly follow the JSON format below, with the analysis content filled into the corresponding fields: \\
\begin{lstlisting}[
  basicstyle=\normalsize\ttfamily,
  breaklines=true,
  frame=none,
  xleftmargin=0.5cm,
  xrightmargin=0.5cm,
  aboveskip=0.5em,
  belowskip=0.5em,
  numbers=none
]
{
  "analysis": "{Detailed analysis of the review process}",
  "check_result": "{true/false}"
}
\end{lstlisting}
\end{tcolorbox}

% The prompt for checking professional terms in PDS
\begin{tcolorbox}[
  colback=mycolor2,
  colframe=mycolor,
  title=\textbf{The prompt for checking professional terms in PDS}, 
  coltitle=black,
  fonttitle=\bfseries\large,
  fontupper=\normalsize\rmfamily,
  breakable
  ]
You are a supervisor for SST psychological counseling. Your task is to review the **counselor's response** to ensure it meets the following criteria and return the result in the specified JSON format:\\

\#\# Counselor's Response\\
\textbf{[counselor response]}\\

\#\# Review Criteria\\
1. Is the language natural and conversational, as in a real counseling dialogue?\\
2. Does it avoid professional or formal expressions (such as \textbf{[term examples]}), especially SST-related technical terms?\\

\#\# Notes\\
1. If all criteria are met, set "check" to true. Otherwise, set it to false.\\
2. If the result is false, **rewrite the response by strictly preserving the original meaning but replacing all professional terms with natural, everyday expressions**. If the result is true, no rewrite is needed.\\

\#\# Output Format\\
Please return the result in the following JSON format:\\
\begin{lstlisting}[
  basicstyle=\normalsize\ttfamily,
  breaklines=true,
  frame=none,
  xleftmargin=0.5cm,
  xrightmargin=0.5cm,
  aboveskip=0.5em,
  belowskip=0.5em,
  numbers=none
]
{
  "check": "{true/false}",
  "feedback": "{analysis of the review process}",
  "rewrite": "{rewritten response}"
}
\end{lstlisting}
\end{tcolorbox}

\newpage

\subsection{Prompts for MDP Synthesis.}
\label{MDP Prompts}

% The prompt for memory capture agent
\begin{tcolorbox}[
  colback=mycolor4,
  colframe=mycolor3,
  title=\textbf{The prompt for memory capture agent}, 
  coltitle=black,
  fonttitle=\bfseries\large,
  fontupper=\normalsize\rmfamily,
  breakable
  ]
\# Task Description \\
You are an SST (Single-Session Therapy) counselor having a session with a client. Your goal in the session is to apply SST therapy knowledge, clarify the client's counseling goal, and help the client find suitable solutions. Based on the dialogue history, complete the following analytical tasks and return the results in the specified JSON format: \\

\#\# Analytical Tasks \\
- Review and summarize the client's personal information. \\
- Analyze and list the client's internal and external resources in detail, and assess their positivity. (If resources have not yet been explored in the dialogue, do not list them.) \\
- Determine whether the counseling goal has been clarified. If so, specify the goal. \\

\#\# Format Specification \\
- Output strictly in the following JSON format, filling in the analysis content in the corresponding fields: \\
\begin{lstlisting}[
  basicstyle=\normalsize\ttfamily,
  breaklines=true,
  frame=none,
  xleftmargin=0.5cm,
  xrightmargin=0.5cm,
  aboveskip=0.5em,
  belowskip=0.5em,
  numbers=none
]
{
  "Personal Information": "{Basic information about the client}",
  "Internal Resources": "{List the client's internal resources and assess their positivity}",
  "External Resources": "{List the client's external resources and assess their positivity}",
  "Counseling Goal": "{The counseling goal}"
}
\end{lstlisting} 

\#\# Dialogue History \\

\textbf{[dialogue history]} \\

\#\# Notes \\
- Internal resources include: the client's past coping strategies, emotional regulation abilities. Positive qualities and past successful coping experiences can be considered positive resources. \\
- External resources include: people around the client (such as family, friends, colleagues), social support systems (such as social institutions, organizations, groups), and external environments (such as work or living environments). \\
- The positivity of a resource depends on whether it helps achieve the counseling goal. \\
- If resources have not yet been explored in the dialogue, write "None" in the corresponding field. \\
- The counseling goal refers to the issue the client wants to solve or the goal they want to achieve in this session. \\
- If the counseling goal has not yet been clarified, write "None" in the corresponding field. \\
\end{tcolorbox}

% The prompt for global plan agent
\begin{tcolorbox}[
  colback=mycolor4,
  colframe=mycolor3,
  title=\textbf{The prompt for global plan agent}, 
  coltitle=black,
  fonttitle=\bfseries\large,
  fontupper=\normalsize\rmfamily,
  breakable
  ]
You are an SST (Single-Session Therapy) counselor conducting a session. To better plan the consultation process, please complete the following reasoning and analysis task based on the *Single‑Session Therapy Manual*, and return the result in the specified JSON format. \\

\#\# Dialogue History \\
\textbf{[dialogue history]} \\

\#\# Counselor's Thoughts \\
\textbf{[counselor thinking]} \\

\#\# Analysis Tasks \\
1. **Determine the current stage and step of the session** \\  
    - **Goal identification Stage (goal)** \\  
      - 1. Building rapport  
      - 2. Clarifying the goal \\  
    - **Working Stage (working)** \\  
      - 1. Exploring positive resources (internal / external) \\  
      - 2. Implementing actions and identifying solutions \\  
    - **Ending Stage (ending)** \\  
2. **Plan the next reply direction based on the consultation goal** \\  
  - Choose **next step**: remain / step shift within stage / stage transition \\  
  - Provide reasoning and suggest the forward direction. \\

\#\# Output Format \\
- Strictly output in the following JSON format, and fill in your detailed analysis in the corresponding fields: \\
\begin{lstlisting}[
  basicstyle=\normalsize\ttfamily,
  breaklines=true,
  frame=none,
  xleftmargin=0.5cm,
  xrightmargin=0.5cm,
  aboveskip=0.5em,
  belowskip=0.5em,
  numbers=none
]
{
  "analysis": "{Your detailed reasoning in response to the analysis tasks}",
  "previous_stage": "{goal/working/ending}",
  "next_step": "{remain/step shift within stage/stage transition}",
  "expected_stage": "{goal/working/ending}"
}
\end{lstlisting} 

\#\# Notes \\

- **remain** is applicable to the following three situations: \\
  1. Currently in the *Goal Identification* stage, at the "Building rapport" step, but the goal has not yet been clarified; \\
  2. Currently in the *Working* stage, at the "Exploring positive resources" step, but resources are not yet fully explored; \\
  3. Currently in the *Working* stage, at the "Implementing actions" step, but still deepening the discussion on how to translate resources into actionable plans. \\
- **step shift within stage** applies only to two transitions: \\
  1. Goal Identification Stage: Building rapport → Clarifying the goal; \\
  2. Working Stage: Exploring positive resources → Implementing actions. \\
- **stage transition** applies only after completing all steps in the current stage, leading to the first step of the next stage. There are only two valid transitions: \\
  1. Goal Identification → Working \\
  2. Working → Ending \\
- The following are *necessary and sufficient* conditions for each valid transition: \\
  1. *Goal Identification → Working*: Counselor and client have **jointly confirmed the consultation goal**. \\
  2. *Working (resources) → Working (actions)*: Resources beneficial to the consultation goal have been identified. \\
  3. *Working (actions) → Ending*: The client has expressed intention to take action and expressed gratitude, with no remaining concerns. \\
\end{tcolorbox}

% The prompt for strategy reasoning agent
\begin{tcolorbox}[
  colback=mycolor4,
  colframe=mycolor3,
  title=\textbf{The prompt for strategy reasoning agent}, 
  coltitle=black,
  fonttitle=\bfseries\large,
  fontupper=\normalsize\rmfamily,
  breakable
  ]
You are an SST counselor conducting a Single-Session Therapy (SST) session. Your core objective is to apply SST knowledge to accurately identify the client’s consultation goal and assist them in finding a suitable solution. Based on the *Single-Session Therapy (SST) Manual*, complete the following analysis task regarding the response, and output the result in the specified JSON format: \\  

\#\# Counselor’s Previous Thoughts \\  
\textbf{[counselor thinking]} \\  

\#\# Response Analysis Task: Based on the client’s current response “\textbf{[client response]}”, conduct the following analytical thinking, and then provide your response content \\  
- According to the planned stage of progression: **\textbf{[expected stage]}**, refer to the SST manual to select an appropriate response strategy for this stage (ensure the strategy matches the response). \\  
- Determine the tone, SST techniques, mode of expression, and response length (the final stage can be moderately extended, other stages must remain concise and clear). \\  

\#\# Format Requirements \\  
- Output strictly in the following JSON format, and fill in the analytical content in the corresponding fields: \\  
\begin{lstlisting}[
  basicstyle=\normalsize\ttfamily,
  breaklines=true,
  frame=none,
  xleftmargin=0.5cm,
  xrightmargin=0.5cm,
  aboveskip=0.5em,
  belowskip=0.5em,
  numbers=none
]
{
  "analysis": "{Analytical thinking behind the response}",
  "response": "{Exact response content}"
}
\end{lstlisting} 

\#\# Notes \\  

* The reply analysis is based on the previous reflection and should focus on choosing an appropriate strategy for the next stage. Do **not** repeat the content from previous thinking; the focus should be on analyzing the client’s current response and planning the next reply. \\  
* Ensure that the **analysis** section does **not** directly state the content of the reply; only the **response** section should present the reply content. \\  
* The counselor’s actual response is “\textbf{[counselor response]}”. The **response** field must match this reply **exactly**, and the method used must align with the strategic thinking in **analysis**. \\  
\end{tcolorbox}

% The prompt for fusion agent
\begin{tcolorbox}[
  colback=mycolor4,
  colframe=mycolor3,
  title=\textbf{The prompt for fusion agent}, 
  coltitle=black,
  fonttitle=\bfseries\large,
  fontupper=\normalsize\rmfamily,
  breakable
]
You are a linguistics expert. Your task is to rewrite the following reflective content by changing all references to the counselor into the first-person pronoun "I". Use appropriate transitional phrases to ensure the logic is clear and coherent. \\

\#\# Reflective Content \\ 
\textbf{[counselor thinking]} \\

\#\# Guidelines \\ 
1. Simulate the counselor’s thinking process in order: first reflect on the *progress of the counseling*, then based on that, analyze the *current stage of counseling*, and finally, based on both, think through the *current session's strategy and response approach*. \\ 
2. Do not alter the original content of each section. Only change the perspective to first-person and add appropriate connectors to create a natural, internally reflective monologue. \\ 
3. The language style should be straightforward, align with the inner voice of “I”, and maintain logical clarity without sudden jumps. Avoid redundancy and ensure the flow is natural. \\ 
4. Do not directly mention any reply content. \\

\#\# Output: Refined Reflection \\ 
...
\end{tcolorbox}

\subsection{Prompts for comparion methods.}
\label{Comparison Prompts}

% The prompt for reasoning models
\begin{tcolorbox}[
  colback=mycolor6,
  colframe=mycolor5,
  title=\textbf{The prompt used for reasoning models}, 
  coltitle=black,
  fonttitle=\bfseries\large,
  fontupper=\normalsize\rmfamily,
  breakable
]
\# SST knowledge \\
\textbf{[SST KB]} \\

You will play the role of a counselor using Single-Session Therapy (SST) in a psychological counseling session, while I will be the client seeking psychological help. Your task is to generate an appropriate response based on the counseling dialogue history and guided by the *Single-Session Therapy (SST) Manual*.
  
First, you will reflect on the dialogue history and the client’s current statement. Then, based on your final reflection, you will respond to the client's current statement. The reflection process should be enclosed in `<think>` and `</think>` tags, for example:
`<think>Content of your reflection</think>Response to the client`.
  
Make sure your response follows the language guidelines below:
  
\#\# Counselor Response Language Guidelines:
  
1. Natural, warm, personable, and conversational—avoid excessive use of professional jargon.
2. The counselor's reply must stay closely connected to what the client just shared.
3. Each reply (except in the closing phase) should not exceed two sentences (less than 50 words).
\end{tcolorbox}

% The prompt for non-reasoning models
\begin{tcolorbox}[
  colback=mycolor6,
  colframe=mycolor5,
  title=\textbf{The prompt used for non-reasoning models}, 
  coltitle=black,
  fonttitle=\bfseries\large,
  fontupper=\normalsize\rmfamily,
  breakable
]
\# SST knowledge \\
\textbf{[SST KB]} \\

You will play the role of a counselor using Single-Session Therapy (SST) in a psychological counseling session, while I will be the client seeking psychological help. Your task is to generate an appropriate response based on the counseling dialogue history and guided by the *Single-Session Therapy (SST) Manual*.
  
Make sure your response follows the language guidelines below:
  
\#\# Counselor Response Language Guidelines:
  
1. Natural, warm, personable, and conversational—avoid excessive use of professional jargon.
2. The counselor's reply must stay closely connected to what the client just shared.
3. Each reply (except in the closing phase) should not exceed two sentences (less than 50 words).
\end{tcolorbox}

\subsection{Prompts for evlaution.}
\label{Evaluation Prompts}

% The prompt for llm-evaluators
\begin{tcolorbox}[
  colback=mycolor6,
  colframe=mycolor5,
  title=\textbf{The prompt used for LLMs to evaluate the dialogue}, 
  coltitle=black,
  fonttitle=\bfseries\large,
  fontupper=\normalsize\rmfamily,
  breakable
]
I want you to act as an evaluator. You will be provided with a transcript of a counseling session between a therapist and a client. Your task is to assess the counselor based on the given criteria. If you believe the therapist falls between two of the descriptors, select the intervening odd number (1, 3, 5). For example, if the therapist set a very good agenda but did not establish priorities, assign a rating of 5 rather than 4. \\

Please follow these steps: \\
1. Read the counseling session transcript carefully. \\
2. Review the evaluation questions and criteria provided below. \\
3. Assign a score based on the criteria, grading very strictly and uptight. If there is any deficiency, no matter how minor, assign a score of 4 or lower. \\
4. Output the score and the explanation, separated by a comma. Do not add any prefix. \\

\#\# Counseling conversation \\
\textbf{[conversation]} \\

\#\# Evaluation Question \\
\textbf{[question]} \\ 

\#\# Scoring Criteria \\
\textbf{[criteria]} \\
\end{tcolorbox}

% The prompt for llm-client
\begin{tcolorbox}[
  colback=mycolor6,
  colframe=mycolor5,
  title=\textbf{The prompt used for LLM client}, 
  coltitle=black,
  fonttitle=\bfseries\large,
  fontupper=\normalsize\rmfamily,
  breakable
]
You are playing the role of a client in a psychological counseling session. Your task is to generate only one suitable response based on the following the counseling dialogue history. \\

\#\# Guidelines for the client's utterance: \\
1. Engage authentically with the counselor's inquiries, reflecting the complexity of emotions and reactions typical in counseling sessions. \\
2. Start the client's utterance with 'Client: '. Ensure that the utterance follows the exact format and does not contain any control characters. \\
3. The client should maintain the following attitude. \\

If you feel that the counseling session has completely ended and meets the end condition, you should include '\textbf{[/END]}' with your utterance. \\
***End Conditions:*** \\
- The client feels that their negative thoughts have been resolved. \\
- The client feels that no further counseling is needed. \\

Please be mindful of these conditions and ensure ****the session should not end prematurely; it must last at least 20 turns.****. \\

\#\# Client Persona and Negative Thoughts: \\
\textbf{[intake form]}\\

\#\# Client's Attitude Towards Counseling: \\
\textbf{[attitude]}\\

Generate only the client's utterance for a single turn and please ensure that your responses do not repeat the client's previous utterances. Do not generate the counselor's part of the dialogue.\\

\#\# Counseling Dialogue History:\\
\textbf{[history]}
\end{tcolorbox}

% The prompt for llm-evaluator Solution Focus evlaution
\begin{tcolorbox}[
  colback=mycolor6,
  colframe=mycolor5,
  title=\textbf{The prompt used for Solution Focus evlaution}, 
  coltitle=black,
  fonttitle=\bfseries\large,
  fontupper=\normalsize\rmfamily,
  breakable
]
\#\# Evaluation Question \\
How effectively does the counselor activate the client's internal and external resources and resilience, and help translate them into usable strategies? \\

\#\# Scoring Criteria \\
Score 0: Ignores or dismisses the client's resources; emphasizes problems or deficiencies, or engages in direct lecturing.\\
Score 2: Mentions some resources or supports, but only at a listing level without exploring their usability or relevance.\\
Score 4: Identifies and specifically explores at least one internal and one external resource, helping the client consider how to apply them to current goals.\\
Score 6: Systematically uncovers a variety of resources, emphasizes past successes and resilience, and integrates the resources into a detailed action plan, significantly enhancing the client’s confidence and motivation.
\end{tcolorbox}

% The prompt for llm-evaluator Resource Activation evlaution
\begin{tcolorbox}[
  colback=mycolor6,
  colframe=mycolor5,
  title=\textbf{The prompt used for Resource Activation evlaution}, 
  coltitle=black,
  fonttitle=\bfseries\large,
  fontupper=\normalsize\rmfamily,
  breakable
]
\#\# Evaluation Question \\
To what extent does the counselor help the client shift from problem narratives toward solutions, hope, and positive experiences? \\

\#\# Scoring Criteria \\
Score 0: The counselor spends most of the time focusing on problem causes or negative emotions; no indication of solution orientation or positive reframing. \\
Score 2: Occasionally brings up possible solutions or positive perspectives, but most exchanges remain focused on problem details; tends to persuade or judge rather than explore. \\
Score 4: For most of the session, the counselor uses techniques such as inquiry, exception-seeking questions, or cost-benefit analysis to guide the client toward solutions, hope, and positive experiences; only a few segments are problem-focused. \\
Score 6: The counselor consistently and flexibly uses solution-focused techniques to help the client autonomously build goals and plans, resulting in a clear emotional or cognitive shift toward the positive; demonstrates a highly balanced combination of empathy and action orientation.
\end{tcolorbox}

% The prompt for llm-evaluator Resource Activation evlaution
\begin{tcolorbox}[
  colback=mycolor6,
  colframe=mycolor5,
  title=\textbf{The prompt used for Goal Orientation evlaution}, 
  coltitle=black,
  fonttitle=\bfseries\large,
  fontupper=\normalsize\rmfamily,
  breakable
]
\#\# Evaluation Question \\
How effectively does the counselor assist the client in setting clear goals, tracking progress, and developing feasible action plans? \\

\#\# Scoring Criteria \\
Score 0: The conversation lacks clear goals; no discussion of actions; the content is unfocused. \\
Score 2: Goals or suggestions are mentioned but remain vague and mostly initiated by the counselor without confirmation from the client; no tracking of progress. \\
Score 4: The counselor collaborates with the client to establish specific, positive, and measurable goals, proposes executable next steps, and revisits the goals or progress at least once during the session. \\
Score 6: Goals are clearly defined early in the session and consistently reviewed and adjusted throughout the early, middle, and late stages; the action plan includes a timeline, resource allocation, and strategies for overcoming obstacles; the client shows strong commitment and actively refines the steps. \\
\end{tcolorbox}

\newpage

\end{document}